\documentclass[10pt,journal,compsoc]{IEEEtran}
\ifCLASSOPTIONcompsoc
  \usepackage[nocompress]{cite}
\else
  \usepackage{cite}
\fi
\ifCLASSINFOpdf
\else
\fi
\usepackage{bbding}

\usepackage{amsfonts}

\usepackage{amsmath}

\usepackage{graphicx}

\usepackage{float}  

\usepackage{subfigure}

\usepackage{bm}

\usepackage{courier}

\usepackage{booktabs}

\usepackage{multirow}

\newcommand{\tabincell}[2]{\begin{tabular}{@{}#1@{}}#2\end{tabular}}

\usepackage{rotating}

\usepackage{amsthm}

\usepackage{footnote}

\usepackage{color}

\usepackage{stfloats}

\usepackage{threeparttable}

\usepackage{adjustbox}

\newtheorem{proposition}{Proposition}

\newtheorem{definition}{Definition}

\newtheorem{property}{Property}

\begin{document}
%
\title{A Principled Design of Image Representation: Towards Forensic Tasks}
%
%
%
%

\author{Shuren~Qi,~Yushu~Zhang,~\IEEEmembership{Member,~IEEE,}~Chao~Wang,\\~Jiantao~Zhou,~\IEEEmembership{Senior~Member,~IEEE},~and~Xiaochun~Cao,~\IEEEmembership{Senior~Member,~IEEE}%
\IEEEcompsocitemizethanks{
	\IEEEcompsocthanksitem S. Qi is with the College of Computer Science and Technology, Nanjing University of Aeronautics and Astronautics, Nanjing, China, and also with the Institute of Information Engineering, Chinese Academy of Sciences, Beijing, China (e-mail: shurenqi@nuaa.edu.cn).
	\IEEEcompsocthanksitem Y. Zhang is with the College of Computer Science and Technology, Nanjing University of Aeronautics and Astronautics, Nanjing, China, and also with the Guangxi Key Laboratory of Trusted Software, Guilin University of Electronic Technology, Guilin, China (e-mail: yushu@nuaa.edu.cn).
	\IEEEcompsocthanksitem C. Wang is with the College of Computer Science and Technology, Nanjing University of Aeronautics and Astronautics, Nanjing, China (e-mail: c.wang@nuaa.edu.cn).
	\IEEEcompsocthanksitem J. Zhou is with the State Key Laboratory of Internet of Things for Smart City, and also with the Department of Computer and Information Science, Faculty of Science and Technology, University of Macau, Macau, China (e-mail: jtzhou@umac.mo). 
	\IEEEcompsocthanksitem X. Cao is with the School of Cyber Science and Technology, Shenzhen Campus of Sun Yat-sen University, Shenzhen, China (e-mail: caoxiaochun@mail.sysu.edu.cn).
	\IEEEcompsocthanksitem Corresponding author: Y. Zhang (e-mail: yushu@nuaa.edu.cn).}
\thanks{IEEE Transactions on Pattern Analysis and Machine Intelligence, 2022, doi: 10.1109/TPAMI.2022.3204971, link: ieeexplore.ieee.org/document/9881995/.}
}

%
%

\markboth{S. Qi \MakeLowercase{\textit{et al.}}: A Principled Design of Image Representation: Towards Forensic Tasks} {S. Qi \MakeLowercase{\textit{et al.}}: A Principled Design of Image Representation: Towards Forensic Tasks}
%



\IEEEtitleabstractindextext{%
\begin{abstract}
Image forensics is a rising topic as the trustworthy multimedia content is critical for modern society. Like other vision-related applications, forensic analysis relies heavily on the proper image representation. Despite the importance, current theoretical understanding for such representation remains limited, with varying degrees of neglect for its key role. For this gap, we attempt to investigate the forensic-oriented image representation as a distinct problem, from the perspectives of theory, implementation, and application. Our work starts from the abstraction of basic principles that the representation for forensics should satisfy, especially revealing the criticality of robustness, interpretability, and coverage. At the theoretical level, we propose a new representation framework for forensics, called \emph{dense invariant representation} (DIR), which is characterized by stable description with mathematical guarantees. At the implementation level, the discrete calculation problems of DIR are discussed, and the corresponding accurate and fast solutions are designed with generic nature and constant complexity. We demonstrate the above arguments on the dense-domain pattern detection and matching experiments, providing comparison results with state-of-the-art descriptors. Also, at the application level, the proposed DIR is initially explored in passive and active forensics, namely copy-move forgery detection and perceptual hashing, exhibiting the benefits in fulfilling the requirements of such forensic tasks.
\end{abstract}

}

\maketitle

\IEEEdisplaynontitleabstractindextext

%
\IEEEpeerreviewmaketitle

\IEEEraisesectionheading{\section{Introduction}\label{sec:intro}}

%
%
%
%

\IEEEPARstart{F}{ake} visual media are now a real-world threat due to their potential abuse to a number of critical scenarios, covering journalism, publishing, judicial investigations, and social networks. For such serious challenges, \emph{image forensics}, aimed at verifying the authenticity and integrity, has attracted widespread attention \cite{ref1}.

Similar to other visual-understanding applications, forensic analysis, i.e., knowledge extraction for the traces of manipulation, relies heavily on proper \emph{image representation}. This is not so surprising, as Herbert Simon asserted: “solving a problem simply means representing it so as to make the solution transparent” \cite{ref2}. Therefore, a common practice in forensic algorithms is to inherit the representation techniques directly from other vision tasks.

We note that, despite the conceptual similarities, image representations for forensic tasks are quite different from the others in the design goals.

Let us begin the analysis from the basic principles that an efficient representation for forensics should satisfy: \emph{Discriminability} – the representation is sufficiently informative to support the distinction between pristine and manipulated data; \emph{Robustness} – the representation is not influenced by geometric or signal perturbations that may be introduced by the adversary; \emph{Coverage} – the representation sufficiently covers the entire image plane, as manipulation may occur in any region including the background; \emph{Interpretability} – the representation should have reliable theoretical guarantees, implying that causation is more important than correlation, due to the role in courts of judicature or in public discussion and debate; \emph{Computational efficiency and accuracy} – the representation should have a reasonable implementation, especially for the widely-used dense processing under constraint of coverage.

\subsection{State of the Art and Motivation}

Starting from the above principles, one can note that the representations in popular vision tasks (e.g., image classification) may not be the best choice for forensic tasks, mainly due to their robustness, interpretability, and coverage properties.

The representations based on \emph{deep learning}, e.g., Convolutional Neural Networks (CNN), have shown impressive performance on a variety of high-level vision tasks. Their success is mainly due to the powerful data-adaptive capability provided by the composition of multiple nonlinear transformations with trainable parameters \cite{ref3}. Despite the strong discriminability, two inherent properties make them inappropriate for many forensic tasks. The first is the difficulty in achieving satisfactory robustness \cite{ref4}, especially for geometric changes \cite{ref5} and adversarial examples \cite{ref6}, which may be used to mislead forensic algorithm by the adversary, i.e., anti-forensics \cite{ref1}. The second is the difficulty in understanding the decision mechanism \cite{ref7}, and such black-box nature reduces the credibility of the forensic results, especially for the critical scenarios like courts \cite{ref1}. We must underline that the research on robustness and interpretability for deep learning is developing \cite{ref8, ref9} but still appears a long way from the requirements of forensic tasks.

Turning now to the topic of hand-crafted feature descriptors. The representations based on \emph{interest points}, e.g., Scale Invariant Feature Transform (SIFT), are very popular and competitive approaches for practical feature engineering. As a distinctive property, keypoint detectors and descriptors are commonly designed to yield a high repeatability under various geometric transformations \cite{ref10}. Such interest points, however, are sparse on the image plane, leading to information loss in some regions. This is clearly detrimental to the tasks that require high coverage, such as fine-grained classification \cite{ref11} and also the forensics [1]. In addition, current research shows that SIFT-like keypoints can be removed or injected through imperceptible pixel modifications \cite{ref12}. This inherent vulnerability has potential applications in the anti-forensics, similar to the adversarial attacks for deep learning.

For these reasons, more forensic researchers have switched to the representation based on \emph{dense sampling}, e.g., DAISY descriptor \cite{ref13}. Such dense representation formed on regular sampling grid, possibly with a range of scales, thus ensuring not only the coverage but also the simple spatial relationships. This regular pattern may be useful in modeling Markov processes, constructing spatial-pooling representations, accelerating feature matching, etc.; while for interest points, the spatial relationship is more arbitrary, resulting in more complicated processing \cite{ref14}. The main challenges, in dense approach, are geometric invariance and implementation efficiency. Existing methods are not stable under geometric distortions, even for common rotation, scaling, and flipping. Moreover, due to the dense nature, the implementation is generally time-consuming, which also limits the use of expensive-but-invariant features.

\subsection{Contributions}

Motivated by above facts, we attempt to present a principled study on the image representation for forensics, covering theory, implementation, and application. To the best of our knowledge, this is a very early work on the theoretical understanding of forensic-oriented image representation, and it was rarely considered as a distinct problem before.

Our main contributions are summarized as follows.

\emph{Theory}. We propose a unified representation framework, named Dense Invariant Representation (DIR), by extending the definition of classical orthogonal moments from the global to the local with scale space. The mathematical analysis explains the important properties of DIR, e.g., the description stability based on covariance. Accordingly, our DIR is able to fulfill the core requirements of forensic tasks, especially for robustness, interpretability, and coverage.

\begin{table}
	\caption{Theoretical Comparison With Related Methods}
	\centering
	\begin{tabular}{ccccc}
		\toprule
		Method & \tabincell{c}{CVPR'08 \\ \cite{ref19}} & \tabincell{c}{TSP'15 \\ \cite{ref20}} & \tabincell{c}{TPAMI'19 \\ \cite{ref21}} & Ours \\ 
		\midrule
		Generic Design &    &    &    & \checkmark \\ 
		Rotation Invariance & \checkmark &    & \checkmark & \checkmark \\ 
		Constant Complexity &    & \checkmark & \checkmark & \checkmark \\ 
		\bottomrule
	\end{tabular}
\end{table}

\emph{Implementation}. We derive an accurate and fast numerical calculation of the DIR. For the accuracy, it is dominated by numerical instability and integration error in the computation of basis functions. The corresponding solutions are recursive strategy and high-precision numerical integration method. For the efficiency, the convolution theorem and scaling theorem of Fourier transform are used to speed up the dense inner products. The resulting complexity does not depend on the window size and thereby is the constant complexity $\mathcal{O}(1)$. Note that such implementation is generic for arbitrary basis functions.

\emph{Application}. We validate the effectiveness of DIR in various simulation experiments and real-world applications. Experiments are performed on dense-domain pattern detection and matching, exhibiting the state-of-the-art performance under challenging geometric transformations or signal corruptions. The direct applications to copy-move forgery detection (passive forensics) and perceptual hashing (active forensics) also demonstrate the accuracy and efficiency gains.

\subsection{Related Works}

We briefly review the topics of dense descriptor and image forensics that are closely related to our work.

\emph{Dense Descriptor}.  In the literature, the designs of dense features is typically based on frequency transform, texture, and orthogonal moments. For the frequency transform, well-known method like Walsh-Hadamard Transform \cite{ref15} and Haar Wavelet \cite{ref16} have been designed for dense pattern matching. Their research focuses on computational complexity and robustness to noise-like attacks, with few considerations on geometric invariance \cite{ref17}. For the texture, the descriptors DAISY \cite{ref13} and DASC \cite{ref18} aim to model the image local structure through gradient distribution and self-similarity, respectively. We regarded them as competitive methods, since their robustness to variable imaging conditions, at the geometric level (e.g., viewing angle) and signal level (e.g., blurring), is highlighted in the application of stereo matching. For orthogonal moments, dense descriptors based on Fourier-Mellin transform \cite{ref19}, Tchebichef moments \cite{ref20}, and Zernike moments \cite{ref21} have appeared in tasks of object detection and image forensics, for their desirable invariance and independence \cite{ref22}. Compared to these theoretically relevant methods, the distinctive properties of our work are summarized in Table 1. As illustrated later, the discussion of DIR on properties and calculations is fully generic to a class of basis functions; also DIR exhibits in-form invariance to rotation and constant complexity to window size.

\emph{Image Forensics}. For the fake content generation, previously, the manipulation was typically performed by Photoshop-like editing software, with the operations such as copy-move \cite{ref23}, splicing \cite{ref24}, and inpainting \cite{ref25}. In recent years, as the rise of deep learning, the so-called deepfake \cite{ref26} allows one to generate realistic fake content in a flexible way. For the fake content detection, the solutions are mainly divided into active and passive forensics. Active ones rely on specific information embedding (e.g., digital watermarking \cite{ref27}) or extracting (e.g., perceptual hashing \cite{ref28}) for the image prior to distribution. Owing to the side information, these methods have desirable property to detect any type of manipulations, but they obviously require additional implementation costs and cannot be used for the images that have been distributed. Passive forensic algorithms, on the contrary, do not rely on such prior processing; they work exclusively on the given image itself. It generally operates by seeking the inconsistencies of given image at the digital \cite{ref29}, physical \cite{ref30}, or semantic \cite{ref31} level, which are inevitably introduced by certain manipulations. On the downside, this line of methods is unsatisfactory in terms of stability and generality, due to the lack of side information. Therefore, it is necessary to design a forensic-oriented image representation for fundamentally improving the accuracy and efficiency of active and passive approaches.

\section{Foundations}

For the sake of completeness, we briefly remind some foundations of classical orthogonal moments (see \cite{ref32} for a survey). In Table 2, we list core notations for this paper.

Mathematically, the general theory of \emph{image moments} is based on the definition of the following inner product $ \left< f,{V_{nm}} \right> $ \cite{ref22}:
\begin{equation}
	\left<f,{V_{nm}} \right> = \iint\limits_D {V_{nm}^*(x,y)f(x,y)dxdy},
\end{equation}
with the image function $f$ and the basis function ${V_{nm}}$ of order $(n,m)\in\mathbb{Z}^2$ on the domain $D\in\mathbb{R}{^2}$, where "$*$" denotes the complex conjugate.

As far as image representation tasks are concerned, two constraints are typically imposed on the explicit forms of basis functions.

\emph{Orthogonality} is a core for achieving various beneficial properties in signal analysis, e.g., information preservation and decorrelation. It occurs when any two basis functions ${V_{nm}}$ and ${V_{n'm'}}$ satisfy the condition:
\begin{equation}
	\begin{split}
		\left<{V_{nm}},{V_{n'm'}}\right> &= \iint\limits_D {{V_{nm}}(x,y)V_{n'm'}^*(x,y)dxdy} \\
		&= {\delta _{nn'}}{\delta _{mm'}},
	\end{split}
\end{equation}
where ${\delta _{\alpha \beta }}$ is the Kronecker delta function: ${\delta _{\alpha \beta } = [\alpha=\beta]}$. For a set of orthogonal functions, the \emph{completeness} in a Hilbert space means its linear span is dense in the space.

\emph{Invariant in form} w.r.t. rotations of axes about the origin $(x,y) = (0,0)$ is a desirable structure in geometric analysis. It means that the basis function ${V_{nm}}$ is expressed in polar coordinates $(r,\theta )$ and is of the form:
\begin{equation}
	{V_{nm}}(\underbrace {r\cos \theta }_x,\underbrace {r\sin \theta }_y) \equiv {V_{nm}}(r,\theta ) = {R_n}(r){A_m}(\theta ),
\end{equation}
with \emph{angular basis function} ${A_m}(\theta ) = \exp (\bm{j}m\theta )$ ($\bm{j} = \sqrt { - 1} $) and \emph{radial basis function} ${R_n}(r)$ could be of any form \cite{ref33}.

By imposing the above two constraints, we now write down a rotation-invariant and orthogonal version of (1) in polar coordinates:
\begin{equation}
	\left< f,{V_{nm}}\right> = \iint\limits_D {R_n^*(r)A_m^*(\theta )f(r,\theta )rdrd\theta },
\end{equation}
where ${R_n}(r)$ should satisfy the weighted orthogonality condition: $\int\limits_0^1 {{R_n}(r)R_{n'}^*(r)rdr}  = \frac{1}{{2\pi }}{\delta _{nn'}}$ and the domain is generally the unit disk: $D = \{ (r,\theta ):r \in [0,1],\theta  \in [0,2\pi )\} $.

In Appendix A, we include some common definitions of ${R_n}(r)$ for (4), with orthogonality and completeness. Note that the main discussion in this paper is generic to all such definitions.

\begin{table}[!t]
	\caption{Notations and Definitions}
	\centering
	\begin{tabular}{cl}
		\toprule
		Notation & Definition \\
		\midrule
		$f$ & The image function \\
		$(x,y)$/$(r,\theta )$/$(i,j)$ & The Cartesian/polar/pixel coordinates \\
		$V$ & The basis function \\
		$(n,m)$/$(u,v)$/$w$ & The order/position/scale parameters \\
		$D$ & The domain of $V$ \\
		$A$/$R$ & The angular/radial basis function \\
		${\cal R}$ & The image representation\\
		${\cal D}$ & The image degradation \\
		${f_{T/R/VF/HF/S}}$ & \tabincell{l}{The translated/rotated/vertical-flipped/\\horizontal-flipped/scaled versions of $f$} \\
		${S_{nm/uv/w}}$ & \tabincell{l}{The sampling sets of $(n,m)$/$(u,v)$/$w$} \\
		${\# _{nm/uv/w}}$ & The sizes of ${S_{nm/uv/w}}$ \\
		$K$ & The constraint on ${S_{nm}}$ \\
		$(N,M)$ & The size of $f$ \\
		$h$ & \tabincell{l}{The integral value of $V$ over a valid pixel region} \\
		$H$ & The kernel w.r.t. $h$ for dense representation \\
		\midrule
		$ \left<\cdot,\cdot\right> $ & The inner product \\
		$*$ & The conjugate of complex \\
		$\angle $ & The phase of complex \\
		$\exp $ & The natural exponential function \\
		$\bm{j}$ & The imaginary unit \\
		${\cal F}$ & The Fourier transform \\
		$T$ & The matrix transpose \\
		$|\cdot|$ & The absolute value \\
		$||\cdot||_p$ & The $p$-norm \\
		$\otimes$ & The convolution \\
		$\odot$ & The point-wise (Hadamard) product \\
		\bottomrule
	\end{tabular}
\end{table}

\section{Dense Invariant Representation: Theory}
This section is dedicated to our continuous-domain representation framework for forensics. Firstly, we explicitly give the mathematical definition of DIR and clarify its theoretical relationship with classical orthogonal moments. Secondly, the properties of DIR regarding geometric transformations are analyzed on the basis of covariance.

\subsection{Definition Extension and Basic Formula}
One important nature of the classical orthogonal moments with form (1) is the \emph{global definition}: the image function $f$ and the basis function ${V_{nm}}$ share the same coordinate system (Cartesian or polar), more precisely, the same origin and similar scale.

Theoretically, such fact renders the classical orthogonal moments useless for local descriptions, and in turn prevents their application in many practical vision problems. For the case of image forensics, especially the forgery localization tasks, a pixel-by-pixel local representation is often a fundamental composition as the coverage requirement. Obviously, the classical definition (1) does not have such a capability.

Here, we aim to meet the coverage requirement in forensic tasks by generalizing the classical definition of image moments. Considering a local coordinate system $(x',y')$ for basis function ${V_{nm}}$, which is a translated and scaled version of the global coordinate system $(x,y)$ with translation offset $(u,v)$ and scale factor $w$. Their explicit relationship is expressed as follows:
\begin{equation}
	 (x',y') = \frac{{(x,y) - (u,v)}}{w},
\end{equation}
and hence the corresponding polar coordinates are:
\begin{equation}
	\left\{ {\begin{array}{*{20}{c}}
		{r' = \sqrt {{{(x')}^2} + {{(y')}^2}}  = \frac{1}{w}\sqrt {{{(x - u)}^2} + {{(y - v)}^2}} }\\
		{\theta ' = \arctan (\frac{{y'}}{{x'}}) = \arctan (\frac{{y - v}}{{x - u}})}
\end{array}} \right..
\end{equation}

\begin{definition}
With the local polar coordinates (6), a local definition of the rotation-invariant and orthogonal moments (4) can be derived as follows:

\begin{small}
\begin{equation}
	\begin{split}
	&\left<f,V_{nm}^{uvw}\right> = \\
	&\iint\limits_D {\underbrace {R_n^*(\overbrace {\frac{{\sqrt {{{(x - u)}^2} + {{(y - v)}^2}} }}{w}}^{r'})A_m^*(\overbrace {\arctan(\frac{{y - u}}{{x - v}})}^{\theta '})}_{{{(V_{nm}^{uvw}(x,y))}^*}}f(x,y)dxdy},
	\end{split}
\end{equation}
\end{small}
where the domain is a disk with center $(u,v)$ and radius $w$ : $D = \{ (x,y):{(x - u)^2} + {(y - v)^2} \le {w^2}\} $.
\end{definition}

\begin{figure}[!t]
	\centering
	\subfigure[]{\includegraphics[scale=0.32]{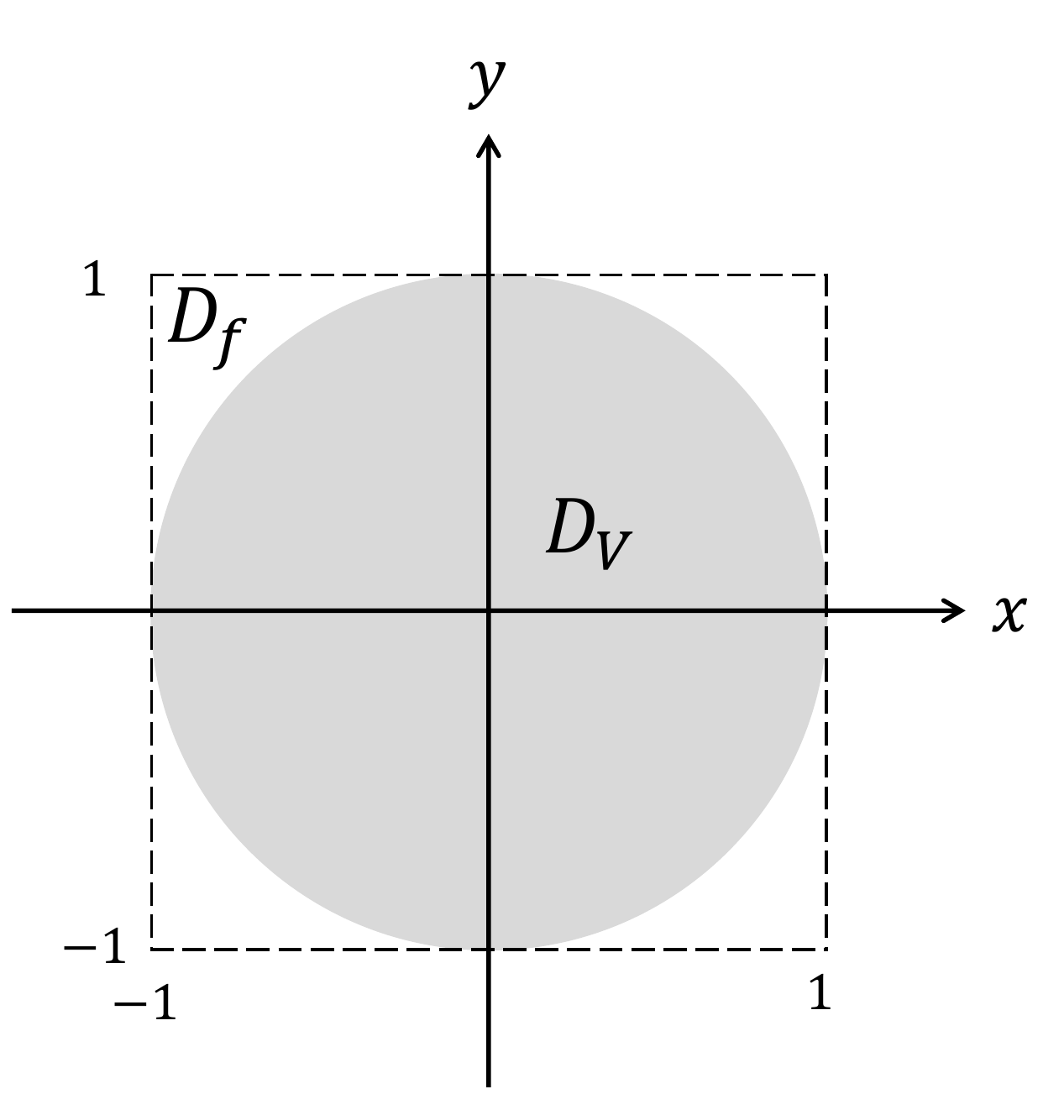}}
	\subfigure[]{\includegraphics[scale=0.32]{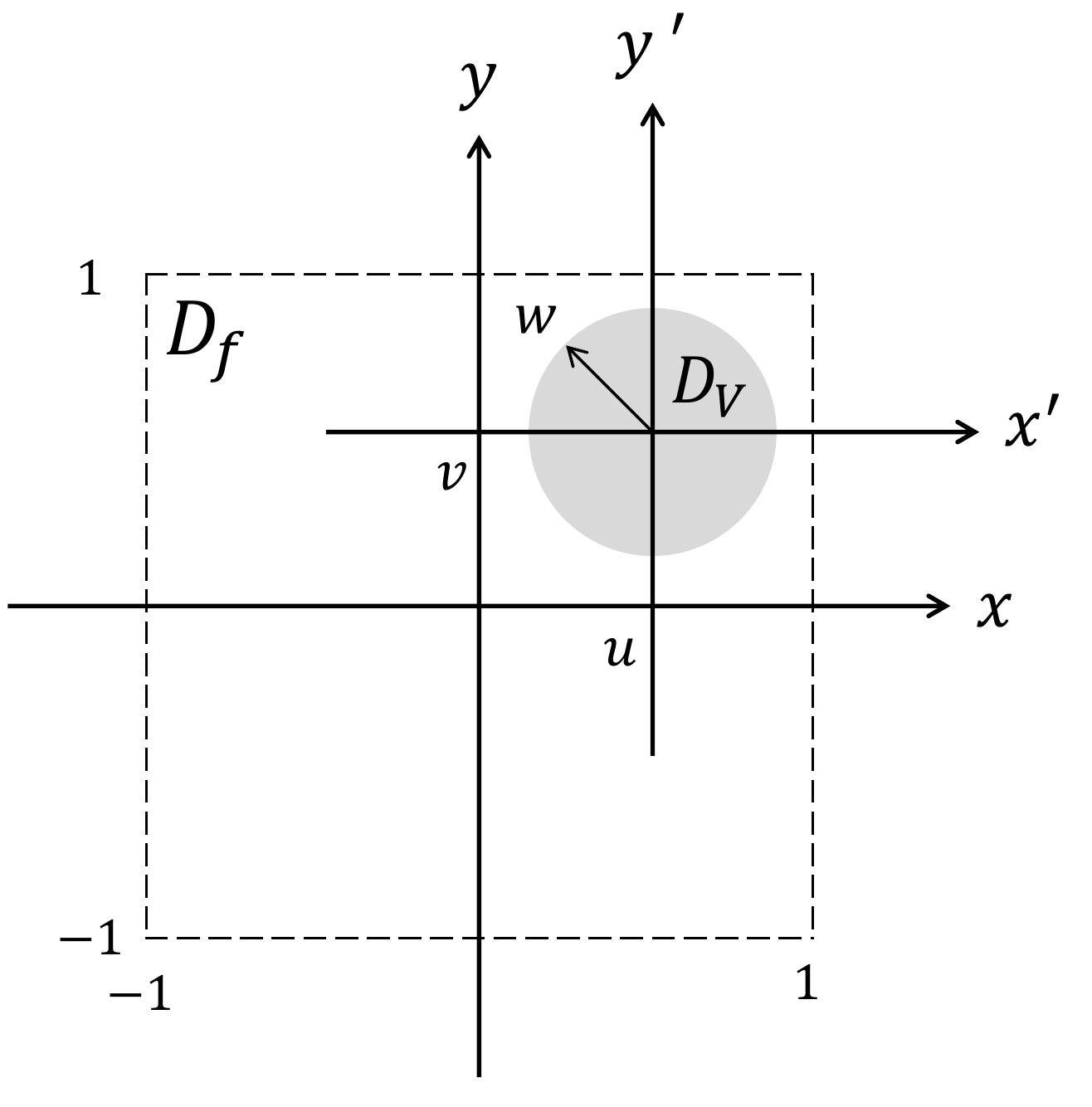}}
	\centering
	\caption{An illustration of the definition extension from the global (a) to the local with scale space (b).}
\end{figure}

The resulting new definition (7) plays a foundational role in constructing the DIR, and therefore it is treated as the \textbf{basic formula} throughout this paper. We would like to make a comment on the notation $V_{nm}^{uvw}$: the superscript and subscript denote the parameters for spatial and frequency domains, respectively. As illustrated in Fig. 1, this new definition allows the domain of image ${D_f}$ and the domain of basis function ${D_V}$ to be built in the separate coordinate systems, providing greater flexibility than the classical orthogonal moments.

Two interesting propositions can be directly observed from this definition, i.e., \emph{generic nature} and \emph{local representation capability}.

\begin{proposition}
The classical form (4) of orthogonal moments is a special case for the basic formula (7), with fixed parameters $(u,v) = (0,0)$ and $w = 1$. 
\end{proposition}

Hence, the basic formula (7) of DIR is regarded as a unified mathematical framework for the research of moments and moment invariants.

\begin{proposition}
With the basic formula (7), the center and scale of local description region $D$ can be controlled by adjusting the parameters $(u,v)$ and $w$, respectively.
\end{proposition}

This local representation capability is also a distinct characteristic of DIR that the classical orthogonal moments do not have, leading to potential use in the local behavior based image processing and visual understanding.

From an analytical perspective, the related descriptors in \cite{ref19, ref21} can be derived from the basic formula (7) by giving a fixed definition of ${R_n}(r)$. This fact means that our approach is a more generic design, and following discussion on the properties and calculations of (7) is also fully generic. As for related work \cite{ref20}, its Cartesian definition of basis functions is inconsistent with (3) and therefore inconsistent with (7). As illustrated later, such difference limits it to achieve the desirable rotation invariance.

\subsection{Important Properties and Representation Formulas}

The representation robustness is quite critical for a range of visual forensic algorithms, due to its intrinsic \emph{two-player} nature \cite{ref1}. Skilled adversary may introduce the trace removal operations such as geometric transformations and signal corruptions to interfere with forensic analysis.

In Section 3.1, a local definition of image moments has been derived, satisfying the coverage requirement. Also, some beneficial properties of classical moment theory, such as completeness and orthogonality, are inherited in (7). However, the local robustness of (7) for geometric transformations has not been investigated, which is clearly distinct from the classical theory based on global assumption.

Next, we aim to meet the robustness requirement in forensic tasks by analyzing the important representation properties under various image transformations.

Our analysis relies on three terms of the representation: \emph{invariance}, \emph{equivariance}, and \emph{covariance} \cite{ref34,ref35}. Considering a representation  ${\cal R}$ and a degradation ${\cal D}$, such terms correspond to the following three identities:
\begin{itemize}
	\item invariance – ${\cal R}({\cal D}(f)) \equiv {\cal R}(f)$,
	\item equivariance – ${\cal R}({\cal D}(f)) \equiv {\cal D}({\cal R}(f))$,
	\item covariance – ${\cal R}({\cal D}(f)) \equiv {\cal D}'({\cal R}(f))$,
\end{itemize}
where ${\cal D}'$ is a composite function of ${\cal D}$. In fact, covariance is a generalized expression of invariance and equivariance. Thus, in general, invariant/equivariant representation can be constructed under the premise that covariance holds.

\begin{figure*}[!t]
	\centering
	\subfigure[]{\includegraphics[scale=0.55]{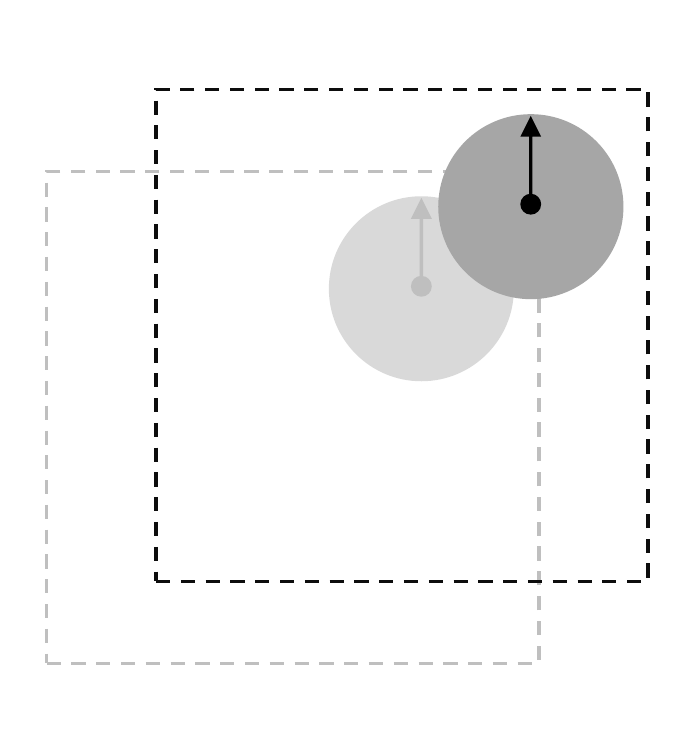}}
	\subfigure[]{\includegraphics[scale=0.55]{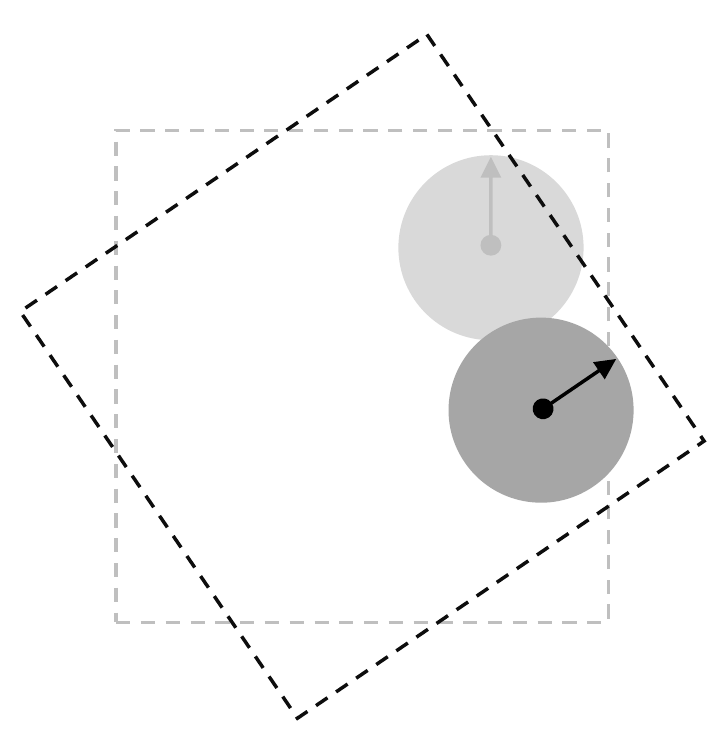}}
	\subfigure[]{\includegraphics[scale=0.55]{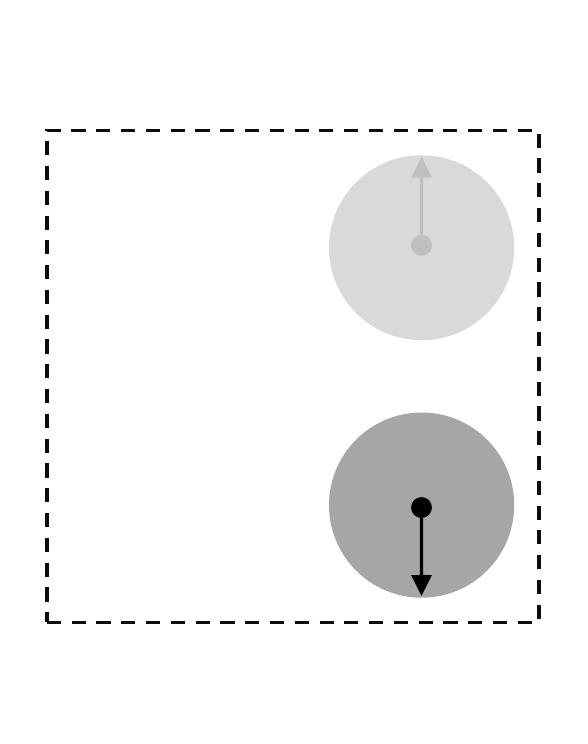}}
	\subfigure[]{\includegraphics[scale=0.55]{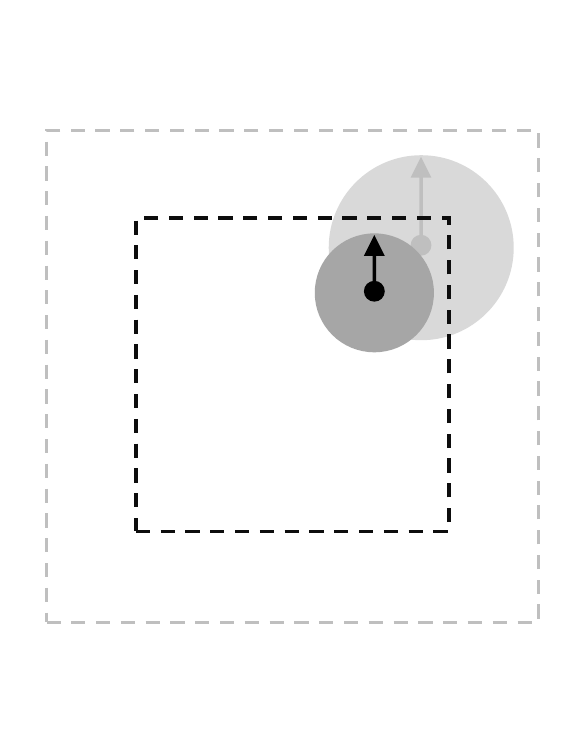}}
	\centering
	\caption{An illustration of the geometric transformations: translation (a), rotation (b), flipping (c), and scaling (d).}
\end{figure*}

\subsubsection{Equivariance to Translation}

Suppose $f_T$ is a translated version of image function $f$ with offset $(\Delta x,\Delta y)$, i.e., ${f_T}(x,y) = f(x + \Delta x,y + \Delta y)$, as shown in Fig. 2. 

\begin{property}
By plugging $f_T$ into (7), it can be checked that image translation operation only affects the representation parameters $(u,v)$, as follows:
\begin{equation}
	\left< {f_T}(x,y),V_{nm}^{uvw}(x,y) \right> 
	= \left< f(x,y),V_{nm}^{(u + \Delta x)(v + \Delta y)w}(x,y) \right>,
\end{equation}
where the same offset $(\Delta x,\Delta y)$ also appears in representation $\left< f(x,y),V_{nm}^{(u + \Delta x)(v + \Delta y)w}(x,y) \right>$, implying the equivariance w.r.t. translation.
\end{property}

\begin{proof}
The proof of (8) is given in Appendix B.
\end{proof}

\begin{proposition}
With the translation equivariance (8), it is feasible to retrieve the translated version of a given pattern over the domain of $(u,v)$ by brute-force feature matching. Moreover, the translation-invariant features can be generated by permutation-invariant mapping, such as average or maximum pooling, over the domain of $(u,v)$. 
\end{proposition}

As seen later, this translation equivariance is also crucial in the derivation of the properties for rotation, flipping, and scaling.

\subsubsection{Invariance to Rotation and Flipping}

Suppose ${f_R}$ is a rotated version of image function $f$ with angle $\phi$ around the center, i.e., ${f_R}(r,\theta ) = f(r,\theta  + \phi )$, where polar coordinates are used for convenience.

Considering any pair of corresponding circular regions in ${f_R}$ and $f$, their geometric relationship can be modeled as a composite of center-aligned rotation and translation, as shown in Fig. 2. Since the translation equivariance has been confirmed, the rest of the analysis will focus only on the center-aligned rotation, i.e., we can restrict the parameters $(u,v) = (0,0)$ without loss of generality.

\begin{property}
By plugging ${f_R}$ into (7) with $(u,v) = (0,0)$ and writing down in polar form, it can be checked that image rotation operation only affects the phase of representation, as follows:
\begin{equation}
	\left< {f_R}(r,\theta),V_{nm}^{uvw}(r',\theta ') \right> = \left< f(r,\theta ),V_{nm}^{uvw}(r',\theta ') \right> A_m^*( - \phi ),
\end{equation}
where the same angle $\phi$ also appears in phase of the representation $\left< f(r,\theta ),V_{nm}^{uvw}(r',\theta ') \right> A_m^*( - \phi )$, implying the covariance w.r.t. rotation.
\end{property}

\begin{proof}
	The proof of (9) is given in Appendix C.
\end{proof}

\begin{proposition}
Considering the covariance (9) of phase, the rotation invariance will hold straightforwardly in the magnitude domain: $| \left< {f_R}(r,\theta),V_{nm}^{uvw}(r',\theta ') \right>|=| \left< f(r,\theta ),V_{nm}^{uvw}(r',\theta ') \right>|$. 
\end{proposition}

Towards rotation invariance, more generalized and information-preserving strategy is based on the proper phase cancelation, which eliminates the effect of the term $A_m^*( - \phi )$. Here, the resulting feature is typically complex-valued containing phase information. In addition to the invariants, retrieving the rotation angle from the phase is easy to solve and robust to noise. For more details on phase cancelation and angle estimation, we address the reader to \cite{ref36}.

\begin{proposition}
By analogy to above derivation, the representations of horizontally flipped version ${f_{HF}}(r,\theta )=f(r,\pi-\theta )$ and vertically one ${f_{VF}}(r,\theta )=f(r,-\theta )$ can be written as ${( - 1)^m}{( \left< f(r,\theta ),V_{nm}^{uvw}(r',\theta ') \right> )^*}$ and ${( \left< f(r,\theta ),V_{nm}^{uvw}(r',\theta ') \right> )^*}$, respectively. Therefore, the flipping invariance will also hold in the magnitude domain.
\end{proposition}

\subsubsection{Covariance to Scaling}

Suppose ${f_S}$ is a scaled version of image function $f$ with factor $s$ around the center, i.e., ${f_S}(x,y) = f(sx,sy)$.

Similarly, the geometric relationship of any pair of corresponding circular regions in ${f_S}$ and $f$ is a composite of center-aligned scaling and translation, as shown in Fig. 2. Taking advantage of translation equivariance, the following derivation is carried out under the setting $(u,v) = (0,0)$.

\begin{property}
By plugging ${f_S}$ into (7) with $(u,v) = (0,0)$, it can be checked that image scaling operation only affects the representation parameters $w$, as follows:
\begin{equation}
	\left< {f_S}(x,y),V_{nm}^{uvw}(x,y) \right> = \left< f(x,y),V_{nm}^{uv(ws)}(x,y) \right>,
\end{equation}
where the same factor $s$ also appears in the representation $\left< f(x,y),V_{nm}^{uv(ws)}(x,y) \right>$, implying the covariance w.r.t. scaling.
\end{property}

\begin{proof}
	The proof of (10) is given in Appendix D.
\end{proof}

\begin{proposition}
With the scaling covariance (10), it is feasible to retrieve the scaled version of a given pattern over the domain of $w$ by brute-force feature matching. Moreover, the scaling-invariant features can be generated by permutation-invariant mapping, such as average or maximum pooling, over the domain of $w$. 
\end{proposition}

We remark that, for the discrete case, this continuous-domain invariance will degenerate to a certain tolerance, as the dense sampling of the scales is difficult in practice. In addition to the permutation-invariant mapping, \emph{scale selection} is also a common path to achieve invariance. Here, the state-of-the-art works in dense scale selection \cite{ref37} are easy to combine with our DIR.

\subsubsection{Orthogonality and Completeness}

Following the classical moment theory, the set of $V_{nm}^{uvw}$ forms an orthogonal and complete basis for Hilbert space, which in turn ensures the uniqueness and independence of the moments \cite{ref22, ref32}. Such information-preservation architecture can thus provide complementary information that improves the discriminability.

Additionally, following the classical moment theory, the representation with low-order $(n,m)$ is more stable under the signal corruptions such as noise, blur, and sampling/quantization effect. The reason is based on the following fact: the lossy signal operations generally work on the high-frequency components of the image, i.e., they mainly affect the high-order moments \cite{ref22, ref32}. In Appendix E, we provide a formal analysis of the robustness to signal corruption from a frequency-domain perspective.

In order to allow the robust low-frequency information preferentially used for the image representation, the set of orders $(n,m)$ is able to constrained by a specific norm and a integer constant $K$:
\begin{equation}
	{S_{nm}}(K) = \{ (n,m):||(n,m)||_p \le K\},
\end{equation}
where $||\cdot||_p$ denotes the $p$-norm and the value of $p$ was typically taken as $1$ or $\infty $ in the literature.

Such properties (8) – (11) provide interpretable guidelines for constructing robust local features, and therefore they are treated as the \textbf{representation formulas} throughout this paper. Note that it is not wise to cover all these invariance types, due to the concerns about discriminability and efficiency, the best way is to take what we need in the application.

In related works \cite{ref19, ref20, ref21}, the properties w.r.t. translation, flipping, scaling, and signal corruptions have hardly been discussed. The \cite{ref19, ref21} exhibit the rotation invariance due to similar covariance in (9); while \cite{ref20} does not have such property as a result of the Cartesian structure.

\section{Dense Invariant Representation: Implementation}

The basic and representation formulas have been described in Sections 3.1 and 3.2, both of them in the continuous domain. Now, we move away from such analytic formulas and concentrate on the discrete implementation of DIR for the digital images. Here, the accuracy and efficiency problems encountered in the discrete implementation of DIR are discussed, and corresponding solutions are provided.

\subsection{Accurate Computation}
The digital-level forensic analysis relies on high-order statistics, e.g., PRNU-like method \cite{ref29}. In such scenarios, computational errors can even dominate the forensic performance. As illustrated later, the direct implementation of DIR is not suitable for the above scenarios.

Next, we aim to meet the computational accuracy requirement in forensic tasks by reducing the approximation error and representation error.

Considering a digital image over the discrete Cartesian grid $\{ f(i,j):(i,j) \in \{ 1,2,...,M\}  \times \{ 1,2,...,N\} \} $, we introduce a continuous version $(x,y) \in [1,M] \times [1,N]$ of discrete variables $(i,j)$ for convenience. Here, a $(i,j)$-centered pixel region is thus defined as ${D_{ij}} = \{ (x,y) \in [i - \frac{{\Delta i}}{2},i + \frac{{\Delta i}}{2}] \times [j - \frac{{\Delta j}}{2},j + \frac{{\Delta j}}{2}]\} $; the value of $f$ over this region is constant and equal to $f(i,j)$.

\begin{definition}
With above notations, the basic formula (7) can be rewritten into discrete form, as follows:
\begin{equation}
	\left< f,V_{nm}^{uvw} \right> = \sum\limits_{(i,j)\mathrm{\;s.t.\;}{D_{ij}} \cap D \ne \emptyset } {h_{nm}^{uvw}(i,j)f(i,j)},
\end{equation}
where $h_{nm}^{uvw}(i,j)$ is the integral value of ${(V_{nm}^{uvw})^*}$ over the intersection of $(i,j)$-centered pixel region ${D_{ij}}$ and the domain of definition $D$, i.e.,
\begin{equation}
	h_{nm}^{uvw}(i,j) = \iint\limits_{{D_{ij}} \cap D}{{{(V_{nm}^{uvw}(x,y))}^*}dxdy}.
\end{equation}
\end{definition}

From a practical perspective, the calculation accuracy of DIR is dominated by (13), due to the continuous integration of complicated functions. More specifically, it normally requires i) determining a numerical integration strategy and ii) calculating the values of the basis functions at the sampling points. Next, we will discuss such two aspects separately.

For the strategy of numerical integration, the \emph{Zero-Order Approximation} (ZOA) is a popular algorithm that directly sets $h_{nm}^{uvw}(i,j) \simeq {(V_{nm}^{uvw}(i,j))^*}\frac{{\Delta i\Delta j}}{w^2}$. Mathematically, its error depends on the frequency of the function, and thus the error may be very significant when the order $(n,m)$ is large or the scale parameter $w$ is small. Note that both cases are common in DIR. Therefore, we suggest that \emph{high-precision numerical integration} methods, e.g., pseudo up-sampling \cite{ref38} and Gaussian quadrature rule \cite{ref39}, should be introduced depending on the application requirements. Such strategies can be uniformly formulated as follows.

\begin{definition}
With the $L$-dimensional cubature formulas, a numerical approximation of the $h_{nm}^{uvw}(i,j)$ in (13) can be derived as:
\begin{equation}
	h_{nm}^{uvw}(i,j) \simeq \sum\limits_{(a,b)} {{c_{ab}}{{(V_{nm}^{uvw}({x_a},{y_b}))}^*}\frac{{\Delta i\Delta j}}{w^2}},
\end{equation}
where $({x_a},{y_b}) \in {D_{ij}}$ is the sampling point with corresponding weight ${c_{ab}}$, and $\#\{(a,b)\} = L$. 
\end{definition}

\begin{proposition}
Regarding the approximation error of (14) with given $(n,m)$, it is in order  $\mathcal{O}((\frac{{\Delta i\Delta j}}{w^2})^{L+1})$. Hence, when larger values of $L$ are used, higher accuracy can be achieved w.r.t. the error order $\mathcal{O}((\frac{{\Delta i\Delta j}}{w^2})^{2})$ of ZOA.
\end{proposition}

For the calculation of basis functions, numerical instability is a common concern, mainly due to the factorial/gamma terms in the Jacobi polynomial based basis functions. Here, we recommend the \emph{recursive strategy} \cite{ref40} to derive the high-order basis functions directly from several low-order ones, without the factorial/gamma of large number.

Here, the discretization (12) – (14) are treated as the \textbf{accurate computation formulas} throughout this paper.

\subsection{Fast Computation}

Starting from the coverage and robustness for forensic analysis, we prefer DIR to work in a dense manner, i.e., the parameters $(u,v)$ and $w$ are sufficiently sampled. In this scenario, the direct computation from definition will exhibit considerable complexity, and thus efficient implementation of DIR is desired in practice.

Next, we aim to meet the computational efficiency requirement in forensic tasks by introducing some useful theorems and data structures. 

\begin{definition}
Considering a dense sampling of $(u,v)$ over the discrete grid $\{ 1,2,...,M\}  \times \{ 1,2,...,N\} $, an equivalent version of (12) can be derived as follows:
\begin{equation}
	\left< f,V_{nm}^{uvw} \right> = f(i,j) \otimes {(H_{nm}^w(i,j))^T},
\end{equation}
where $\otimes$ denotes the convolution operation and ${( \cdot )^T}$ indicates the matrix transpose; $H_{nm}^w$ is a kernel defined by $h_{nm}^{uvw}$ with following form:
\begin{equation}
	H_{nm}^w(i,j) = \{ h_{nm}^{uvw}(i,j):u,v = w,(i,j)\mathrm{\;s.t.\;}{D_{ij}} \cap D \ne \emptyset \}.
\end{equation}
\end{definition}

Here, the dense inner product in (12) is converted to the convolution in (15). Note that we use “convolution” instead of the similar term “cross-correlation” is to facilitate subsequent discussion.

Regarding the computational complexity of (15) with given $(n,m)$, it is determined by the size of kernel $H_{nm}^w$ (i.e., $4{w^2}$), size of sample set ${S_{uv}}$ for parameter $(u,v)$ (denoted as $\#_{uv}$), and size of sample set ${S_w}$ for parameter $w$ (denoted as $\#_{w}$); hence a total of $\Theta (4\sum\limits_{w \in {S_w}} {{w^2}\#_{uv}} )$ or $\mathcal{O}({w_{\max }}^2\#_{uv}\#_{w})$ multiplications, where ${w_{\max }}$ is the maximum in ${S_w}$. As for (16), the complexity is almost negligible, because the kernel only needs to calculate once for given $n$, $m$ and $w$, without dense processing.

Now, we show that there are efficient ways to conduct the DIR.
\begin{definition}
	Let us introduce the \emph{convolution theorem} of the Fourier transform, such that the spatial-domain convolution in (15) can be converted to a frequency-domain product as \cite{ref41}:
\begin{equation}
	 \left< f,V_{nm}^{uvw} \right> = {\mathcal{F}^{ - 1}}(\mathcal{F}(f) \odot \mathcal{F}({(H_{nm}^w)^T})),
\end{equation}
where $\mathcal{F}$ denotes the Fourier transform and $\odot$ indicates the point-wise multiplication.
\end{definition}

\begin{proposition}
If the Fast Fourier Transform (FFT) algorithm is used for the implementation of (17), the multiplication complexity will become $\Theta (\#_{w}(3\#_{uv}\log\#_{uv} + \#_{uv}))$ or $\mathcal{O}(\#_{w}\#_{uv}\log\#_{uv})$.
\end{proposition}

It should be highlighted that the scale parameter $w$ has no role in the complexity, meaning a constant-time calculation w.r.t. the kernel/window size. Ideally, when ${w_{\max }}$ is large enough such that ${w_{\max }}^2$ is greater than $\log (\#_{uv})$, this FFT-based method will take less time than the convolution-based one in (15). This observation is crucial due to the fact that the sampling of $w$ generally involves some large values, for covering the scale variations in the application.
 
Further, we point out where the above FFT-based algorithm can be accelerated. 

\begin{definition}
Let us introduce the \emph{scaling theorem} of the Fourier transform, allowing the frequency coefficients $\mathcal{F}({(H_{nm}^w)^T})$ to be derived directly from the calculated ones with scale parameter ${w_0}$:
\begin{equation}
	\mathcal{F}({(H_{nm}^w)^T}) = {(\frac{w}{{{w_0}}})^2}\mathcal{F}({(H_{nm}^{{w_0}})^T})(\frac{{w{\xi _i}}}{{{w_0}}},\frac{{w{\xi _j}}}{{{w_0}}}),
\end{equation}
where $({\xi _i},{\xi _j})$ are frequency variables.
\end{definition}

\begin{proposition}
In practice, therefore, for given $(n,m)$ and a pair of $w$ and ${w_0}$, the direct FFT can be replaced with the computationally inexpensive interpolation as (18), ideally saving $\Theta (\#_{uv}\log\#_{uv})$ multiplications.
\end{proposition}

In addition to convolution and scaling theorems, we also note an implementation trick for some practical scenarios.

\begin{proposition}
Going back to (17), a valuable observation is that the term $\mathcal{F}({(H_{nm}^w)^T})$ is independent of the content of the input $f$ and is only relevant to its size. Thus, a \emph{lookup table} with indexes $(n,m,w)$ can be pre-calculated, which contains the ready-to-use terms $\mathcal{F}({(H_{nm}^w)^T})$ of fixed size ${M_0} \times {N_0}$. Also, the size of the digital image $f$ will be normalized to ${M_0} \times {N_0}$. Therefore, for given $(n,m)$ and $w$, the lookup table strategy ideally saves $\Theta (\#_{uv}\log \#_{uv})$ multiplications.
\end{proposition}

Here, the implementation (15) – (18) are treated as the \textbf{fast computation formulas} throughout this paper. 

In related works \cite{ref19, ref20, ref21}, the accurate computation has hardly been discussed. The \cite{ref20, ref21} derive fast computation strategy with the constant complexity w.r.t. kernel/window size, but both rely heavily on specific basis functions. In contrast, our constant-time calculation is built on a generic framework, regardless of the specific definition of basis functions.

It is worth to remark that accurate strategy (14) is mainly for small $w$, while the fast strategy (17) is more beneficial with large $w$.

\begin{figure*}[!t]
	\centering
	\includegraphics[scale=0.65]{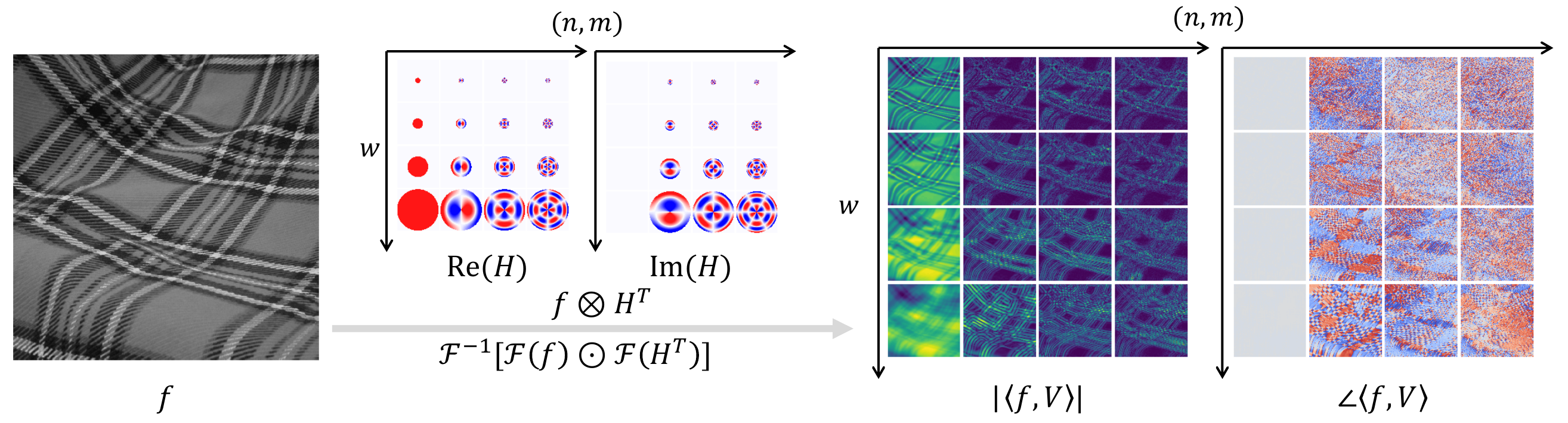}
	\caption{A high-level intuition of the proposed dense invariant representation. }
\end{figure*}

\section{Practical Intuitions and Guidelines}

In this section, we answer several practical questions that could be raised in the applications of DIR.

\subsection{Practical Intuitions}
For the practical intuitions of Sections 3 and 4, we illustrate the DIR by considering the analysis of a texture image, as shown in Fig. 3 \footnote{The real/imaginary parts of DIR kernels and the phases of DIR coefficients are displayed in a colormap ranging from blue to white to red, where 0 corresponding to the white; the magnitudes of DIR coefficients are displayed in a colormap ranging from black to yellow, where 0 corresponding to the black.}.

\emph{Computation}. Going back to (13), (14), and (16), one can estimate a set of kernels $H_{nm}^w$, which are derived from the basis functions $V_{nm}^{uvw}$ with different orders $(n,m)$ and scales $w$. The real and imaginary parts of $H_{nm}^w$ are given in Fig. 3. Then, by applying spatial-domain convolution (15) of image $f$ with the transpose of such kernels $H_{nm}^w$, or its equivalent frequency-domain counterpart (17), the image $f$ is decomposed into dense coefficients $\left< f,V_{nm}^{uvw} \right>$ along multiple orders and scales.

\emph{Representation}. As illustrated in Fig. 3, the magnitudes of such coefficients capture the main structural information (e.g., the line patterns), while the phases provide very detailed properties (e.g., the rich textures). For robust representation, the invariants can be simply built by magnitude (w.r.t. rotation and flipping) and pooling (w.r.t. translation and scaling) functions over the coefficients, referring to properties (8) – (11). Typically, this magnitude-only strategy is sufficient in practice. As for more informative representation, the expressive phases can be further encoded into the features, based on appropriate phase cancellation.

\subsection{Practical Guidelines}

Here, we give some general principles for setting parameters and radial basis functions, which support the applications of DIR for practical forensic problems.

Considering the clear physical meaning of DIR parameters $(n,m)$, $(u,v)$, and $w$, their settings can be directly derived from the mathematical properties of the forensic problem.

\begin{itemize}
	\item Regarding the $(n,m)$, it controls the \emph{frequency} of the represented image information. For the forensics that reveal the semantic/physical/digital artifacts w.r.t. low/mid/high-frequency information (following the taxonomy of \cite{ref1}), the pertinent forensic analysis can be achieved by low/mid/high-order DIR.
	\item Regarding the $(u,v)$, it controls the \emph{position} of the represented image information. In general, the dense sampling is a good choice for avoiding false negatives (under the coverage principle).  In practice, such dense strategy may sometimes be compromised to interval sampling for reducing complexity.
	\item Regarding the $w$, it controls the \emph{scale} of the represented image information. For a given forensic task that does not require scale invariance, $w$ can be directly taken as a fixed value. While for the task that requires such invariance, $w$ should be sampled sufficiently. Note that if the prior knowledge w.r.t. the scale of the pattern under analysis, such knowledge should be introduced in the setting to avoid unnecessary sampling.
\end{itemize}

As for the definition of radial basis function in DIR (as listed in Appendix A), the choice mainly refers to the representation capability and computational accuracy/efficiency.

\begin{itemize}
	\item Regarding the representation capability, it is mainly influenced by the distribution of the zeros of radial basis function. Theoretically, a uniform distribution of zeros on $[0,1]$ is optimal. Another alternative path is to combine radial basis functions with complementary distributions of zeros for capturing complementary image information \cite{ref40}.
	\item Regarding the computational accuracy/efficiency, it is mainly influenced by the basic mathematical properties of radial basis function. Here, the complexity is related to whether factorial/gamma terms, summation/series operations, and root-finding processes are involved. The numerical stability is related to whether the factorial/gamma terms and very high absolute values are involved.
\end{itemize}

For the radial basis function, interested readers can access relevant knowledge from our survey paper \cite{ref32}.
	
We also would like to clarify that when the DIR settings satisfy such general principles, the performance gap between the different settings is generally acceptable (see Appendixes F and G), implying that the DIR is not sensitive to such settings.

\begin{figure*}[!t]
	\centering
	\subfigure[]{\includegraphics[scale=0.9]{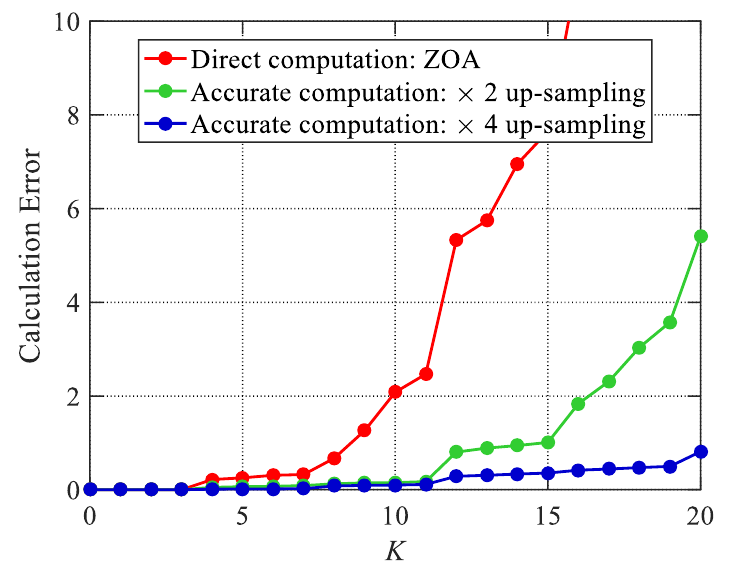}}
	\subfigure[]{\includegraphics[scale=0.9]{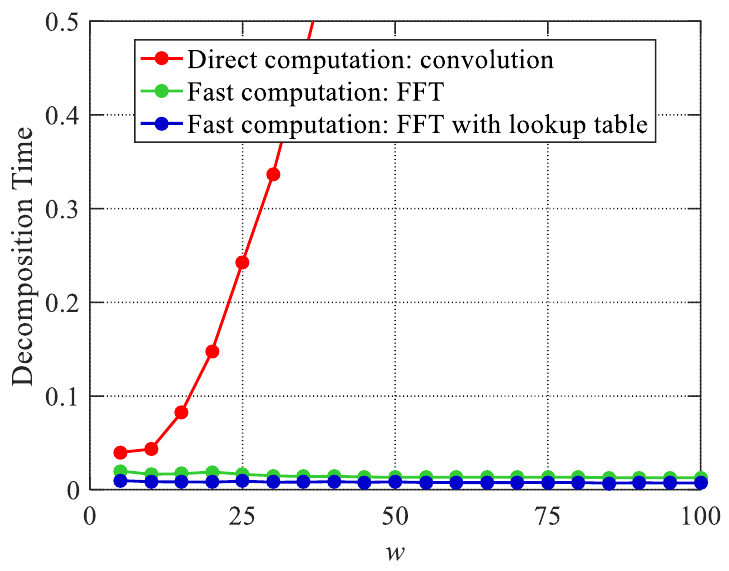}}
	\centering
	\caption{Calculation error (a) and decomposition time (b) for different implementation methods.}
\end{figure*}

\section{Experiments}

In this section, we will evaluate the performance of the proposed DIR, covering experiments at both the descriptor level and the forensic level.

At the descriptor level, the performance statistics for accuracy/complexity and robustness/invariance of DIR are provided. For the accuracy and complexity, several implementation strategies in Section 4 will be compared quantitatively. For the robustness and invariance, our DIR and state-of-the-art dense descriptors are tested in dense detection and matching tasks under geometric or signal perturbations.

At the forensic level, we will measure the usefulness of the proposed DIR in passive and active semantic forensic paths, where the copy-move forgery detection and perceptual hashing are considered as examples, respectively. Here, the DIR is applied directly to the algorithm as a feature extraction module. Such a direct application will be compared with state-of-the-art methods in terms of accuracy and efficiency. In Appendixes F and G, we provide some additional forensic experiments; in Appendix H, we discuss the applicability of DIR to forensics beyond exposing semantic artifacts.

For the sake of brevity, a set of cosine functions is chosen as the radial basis functions in all experiments, i.e., a DIR extended from the Polar Cosine Transform (PCT) \cite{ref42}. In fact, we implement all the DIR extensions for the methods listed in Appendix A. The code is available online at \texttt{https://github.com/ShurenQi/DIR}. Note that all experiments are performed in Matlab R2021a, with 2.90-GHz CPU and 16-GB RAM, under Microsoft Windows environment.

\subsection{Accuracy and Complexity}

\emph{Accuracy}. Firstly, we evaluate the accuracy performance of different implementation strategies in Section 4.1. Let us consider an image with unity gray-level $\{ {f_{{\rm{uni}}}}(x,y) = 1:(x,y) \in D\} $, such that the moments with form (7) can be easily derived as:
\begin{equation}
	\left< {f_{{\rm{uni}}}},V_{nm}^{uvw} \right> = \left\{ {\begin{array}{*{20}{c}}
			0&{m \ne 0}\\
			{2\pi \int\limits_0^1 {R_n^*(r')r'dr'} }&{m = 0}
	\end{array}} \right..
\end{equation}

This equation indicates that the theoretical value of $\left< {f_{{\rm{uni}}}},V_{nm}^{uvw} \right> $ is zero for $m \ne 0$ case, but in practice such property usually does not hold due to various implementation errors. Hence, we can introduce a simple measure for evaluating the error, named Calculation Error (CE), as follows \cite{ref32}:
\begin{equation}
	{\rm{CE}} = \sum\limits_{(n,m) \in \{ {\bf{S}}(K),m \ne 0\} } {|\widehat { \left< {f_{{\rm{uni}}}},V_{nm}^{uvw} \right> }|},
\end{equation}
where $\widehat { \left< {f_{{\rm{uni}}}},V_{nm}^{uvw} \right> }$ is the actual calculated value by a discrete implementation.

The comparison methods include the direct computation by ZOA and the accurate computation by pseudo up-sampling (14). As we mentioned, in theory, the error is positively/inversely proportional to the frequency/number of samples, thus we consider $w = 8$ and $K \in \{ 0,1,...,20\} $.

The comparison results of CE are shown in Fig. 4 (a). We observe that the up-sampling strategy yields less error than the ZOA strategy. This observation is especially true for higher up-sampling rate and higher order constraint $K$. Accordingly, the coefficient with small $w$ and large $(n,m)$ by direct computation may contains considerable errors, hence compromising the discriminability. Such experimental evidence supports our theoretical analysis on integration error, verifying the usefulness of the accurate strategy (14) for real-world scenarios.

\emph{Complexity}. Secondly, we evaluate the complexity performance of different implementation strategies in Section 4.2. The experiment is designed to compute the dense coefficients for an image of size $512 \times 512$, i.e., $(u,v) \in {\{ 1,2,...,512\} ^2}$, under fixed order $(n,m) = (1,1)$ and varying scales $w \in \{ 5,10,...,200\} $. The measure is the Decomposition Time (DT), i.e., the CPU elapsed time in single-thread modality. The comparison methods include the direct computation by convolution (15) and the fast computation by FFT (17), where the trick of lookup table is also considered.

The comparison results of DT are shown in Fig. 4 (b). As $w$ increases, the DT curve for the convolution-based computation rises sharply, while the FFT-based computations exhibit almost-constant time costs. Obviously, this observation is consistent with our theoretical expectation on the complexity. In addition, the introduction of the lookup table further reduces the running time of FFT-based computation. Note that only the computation time for a given $(n,m,w)$ is considered here. While in real-world scenarios, since a set of $(n,m,w)$ is generally needed, the performance gap between the two strategies will be even greater.

\begin{figure*}[!t]
	\centering
	\includegraphics[scale=0.8]{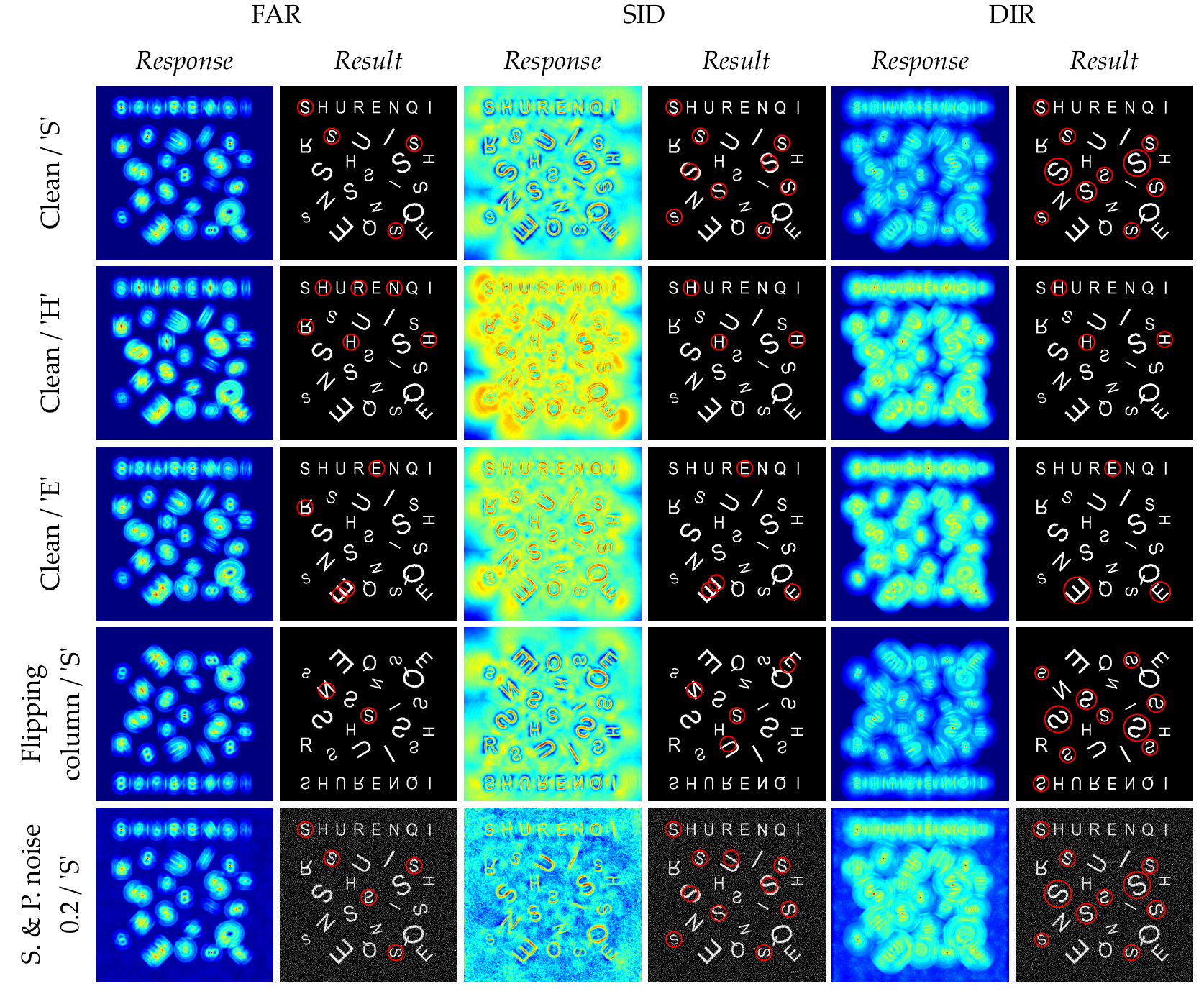}
	\centering
	\caption{Some samples of the pattern detection by different dense descriptors.}
\end{figure*}

\subsection{Robustness and Invariance}

Two typical vision tasks, dense pattern detection and matching, are chosen for evaluating the robustness and invariance. Note that such two tasks have different natures in the representation: the detection generally works with large window, allowing the use of multi-scale representation as the results are sparse; while the matching usually relies on small window, forcing the use of compact representation as the results are dense.

In the experiments, the algorithm framework is designed in a simple form. This is because we aim to reflect the inherent performance of different descriptors in the framework, rather than directly achieving state-of-the-art results on such tasks.

\begin{figure*}[!t]
	\centering
	\includegraphics[scale=0.8]{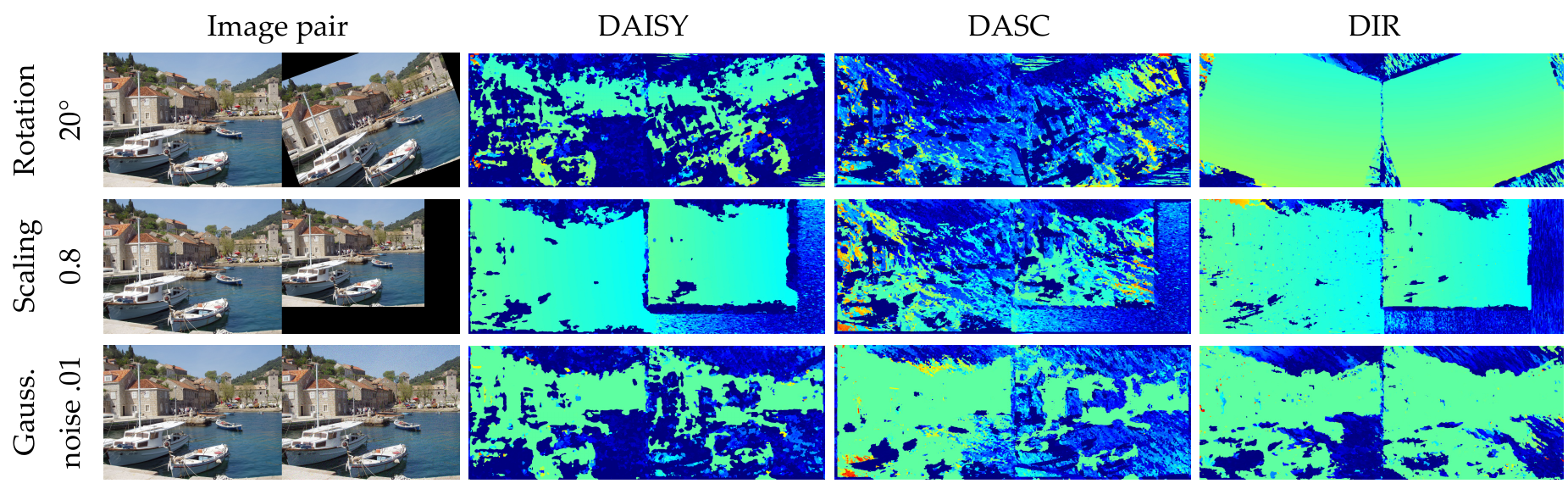}
	\centering
	\caption{Some samples of the pattern matching by different dense descriptors.}
\end{figure*}

\subsubsection{Dense Detection}

For this experiment, we consider detecting five letter ‘S’, ‘H’, ‘R’, ‘E’, and ‘N’ from a $800 \times 800$ image containing multiple letters and their rotated/scaled/flipped versions, as shown in Fig. 5. Note that other letters are not involved as their detection leads to false positives in backgrounds or lines. Furthermore, this image is globally degraded with different affine transformations and signal corruptions, leading to more challenging conditions.

\emph{Algorithm Design}. The template is a patch taken directly from the top left corner of the image, containing a clean letter ‘S’ with the fixed orientation and size. The calculated feature vector of this template is considered to be the ground-truth for comparing with the dense features from the host image. By calculating the Euclidean distances of such features, we treat the positions with smaller distance values as the pattern detection results. We use the F1 score to evaluate the detection performance.

The competing methods include the state-of-the-art invariant descriptors Fourier-Argand Representation (FAR) \cite{ref43} and Scale Invariant Descriptor (SID) \cite{ref19}. Here, the FAR is designed for rotation-invariant pattern detection, with impressive robustness to severe noise conditions. The SID is characterized by the inherent rotation and scaling invariance, without the need for common pooling or selection operations.

For implementation details of DIR, we set the order constraint $K = 5$ in (11) with $\infty $-norm, resulting in 36-dimensional features for each position $(u,v,w)$ in the scale space. Here, we consider a dense sampling of $(u,v)$ over the image grid, and the scale $w$ increases from 30 to 120 with 10 samples. As for FAR and SID, we use the common parameter settings in the original papers. Note that both FAR and SID work on the single scale, and their feature dimensions at each position are 21 and 1008, respectively. In terms of sampling, the FAR is dense, while the SID employs a 2-pixel sampling interval due to the considerable time/space cost.

\begin{table}
	\caption{F1 Scores (\%) for Different Dense Descriptors in Pattern Detection Experiment.}
	\centering
	\begin{tabular}{cccc}
		\toprule
		Method & \tabincell{c}{FAR \\ \cite{ref43}} & \tabincell{c}{SID \\ \cite{ref19}} & DIR \\
		\midrule
		Clean                      & 61.54 & 92.68 & 97.56  \\
		\midrule
		Rotation 20°               & 60.00 & 73.08 & 100.00 \\
		Rotation 45°               & 56.41 & 56.25 & 97.56  \\
		Flipping column            & 34.48 & 31.58 & 97.56  \\
		Flipping row               & 34.48 & 26.47 & 97.56  \\
		Scaling 0.8                & 25.64 & 35.05 & 74.07  \\
		Scaling 0.5                & 19.35 & 19.18 & 45.61  \\
		\midrule
		Gaussian noise 0.01        & 61.54 & 92.31 & 100.00 \\
		Gaussian noise 0.02        & 61.54 & 92.68 & 93.33  \\
		Salt and pepper noise 0.01 & 60.00 & 90.48 & 97.56  \\
		Salt and pepper noise 0.02 & 60.00 & 92.68 & 100.00 \\
		Average filtering 7 × 7    & 43.24 & 45.57 & 61.76  \\
		Average filtering 9 × 9    & 40.00 & 37.11 & 60.61  \\
		Gaussian filtering 7 × 7   & 45.71 & 50.75 & 84.00  \\
		Gaussian filtering 9 × 9   & 43.24 & 46.34 & 59.15  \\
		Median filtering 7 × 7     & 60.00 & 92.68 & 100.00 \\
		Median filtering 9 × 9     & 61.22 & 92.68 & 97.67  \\
		JPEG compression 10        & 61.54 & 92.68 & 100.00 \\
		JPEG compression 5         & 61.54 & 92.68 & 100.00 \\
		Laplacian sharpening       & 57.89 & 92.68 & 100.00 \\
		\midrule
		\emph{Average} $\uparrow$ & 50.47 & 67.28 & 88.20  \\
		\emph{Standard deviation} $\downarrow$ & 13.26 & 27.26  & 17.12\\
		\bottomrule
	\end{tabular}
\end{table}

\emph{Robustness and Invariance}. Fig. 5 shows some samples of the response map and detection result, while the F1 scores for all above comparison methods are given in Table 3. As can be observed here, under the degradation operations, it is challenging for the descriptor to be robust while maintaining discriminability. The FAR exhibits rotation invariance and higher tolerance for severe noise than other methods. However, the FAR cannot handle scale changes, and the features are relatively unstable under the filtering-like operations. As for the SID, its log-polar sampling allows the rotation/scaling-invariant representation on the single scale. However, the response maps suggest that the features are weak in rejecting the irrelevant letters i.e., less discriminability. In addition, both FAR and SID are not flip-invariant. In contrast, the proposed DIR exhibits stronger stability for rotation, flipping and scaling, which should be attributed to the full exploitation of the covariance. Also, thanks to the orthogonality and completeness of the basis functions, our DIR is able to distinguish well between relevant and irrelevant patterns under signal corruptions.

\emph{Efficiency}. In terms of efficiency, the DIR maintains a reasonable time cost: $ \sim $ 2 seconds, $ \sim $ 35 seconds, and $ \sim $ 6 seconds for FAR, SID, and DIR, respectively, even though our method works on multiple scales.

\begin{table}
	\caption{Repeatability Scores  (\%) for Different Dense Descriptors in Pattern Matching Experiment.}
	\centering
	\begin{tabular}{cccc}
		\toprule
		Method & \tabincell{c}{DAISY \\ \cite{ref13}} & \tabincell{c}{DASC \\ \cite{ref18}} & DIR \\ 
		\midrule
		Clean & 95.58 & 94.48 & 94.26 \\ 
		\midrule
		Rotation 20° & 28.67 & 0.16 & 85.01 \\ 
		Rotation 45° & 0.30 & 0.35 & 76.61 \\ 
		Flipping column & 0.23 & 0.15 & 92.81 \\ 
		Flipping row & 0.48 & 0.13 & 91.03 \\ 
		Scaling 0.8 & 72.07 & 7.54 & 49.36 \\ 
		Scaling 1.3 & 74.79 & 3.20 & 32.27 \\ 
		\midrule
		Gaussian noise 0.01 & 23.66 & 31.43 & 35.60 \\ 
		Gaussian noise 0.02 & 15.54 & 20.73 & 21.98 \\ 
		Salt and pepper noise 0.01 & 60.17 & 63.68 & 58.96 \\ 
		Salt and pepper noise 0.02 & 45.44 & 47.59 & 45.43 \\ 
		Average filtering 5 × 5 & 52.56 & 68.09 & 60.53 \\ 
		Average filtering 7 × 7 & 27.58 & 29.63 & 18.77 \\ 
		Gaussian filtering 5 × 5 & 54.81 & 69.70 & 64.47 \\
		Gaussian filtering 7 × 7 & 32.87 & 41.91 & 29.95 \\ 
		Median filtering 5 × 5 & 54.09 & 72.93 & 59.88 \\ 
		Median filtering 7 × 7 & 28.84 & 42.30 & 32.21 \\ 
		JPEG compression 10 & 48.49 & 40.25 & 35.97 \\ 
		JPEG compression 5 & 28.15 & 17.02 & 13.69 \\ 
		Laplacian sharpening & 43.75 & 79.58 & 47.39 \\ 
		\midrule
		\emph{Average} $\uparrow$ & 39.40  & 36.54  & 52.31    \\
		\emph{Standard deviation} $\downarrow$ & 25.10  & 29.58  & 24.84  \\
		\bottomrule
	\end{tabular}
\end{table}

\subsubsection{Dense Matching}

For this experiment, we consider establishing dense correspondences between original image and its degraded version, as shown in Fig. 6. Here, 10 images from the INRIA Holidays dataset \cite{ref73} are selected for providing average experimental results, and such images are normalized to a same size of 1000 × 1333.

\emph{Algorithm Design}. The experiment is performed on a common framework: the PatchMatch \cite{ref74} for matching the dense features, and the RANSAC \cite{ref75} for excluding false matches through data modeling. Note that the regular spatial relationship of dense features allows PatchMatch to achieve geometric-invariant matching with high efficiency \cite{ref76}. We use the repeatability score \cite{ref10} to evaluate the matching performance.

The competing methods include the state-of-the-art dense descriptors DAISY \cite{ref13} and DASC \cite{ref18}. Here, the DAISY is designed for wide-baseline stereo matching, hence considering large perspective distortions. The DASC further developed the idea of DAISY, paying special attention to photometric variations.

For implementation details of DIR, we chose a set of $(n,m)$ with a maximum order of 3, generating 10-dimensional features for each position $(u,v,w)$ in the scale space. Here, we consider a dense sampling of $(u,v)$ over the image grid, and the scale $w$ increases from 8 to 32 with 10 samples. For a compact representation with scaling tolerance, the feature vectors are average-pooled together over the scales. As for DAISY and DASC, we use the common parameter settings in the original papers. Note that feature dimensions of DAISY and DASC at each position are 200 and 128, respectively, much more than the 10 dimensions of DIR.

\begin{figure*}[!t]
	\centering
	\includegraphics[scale=0.57]{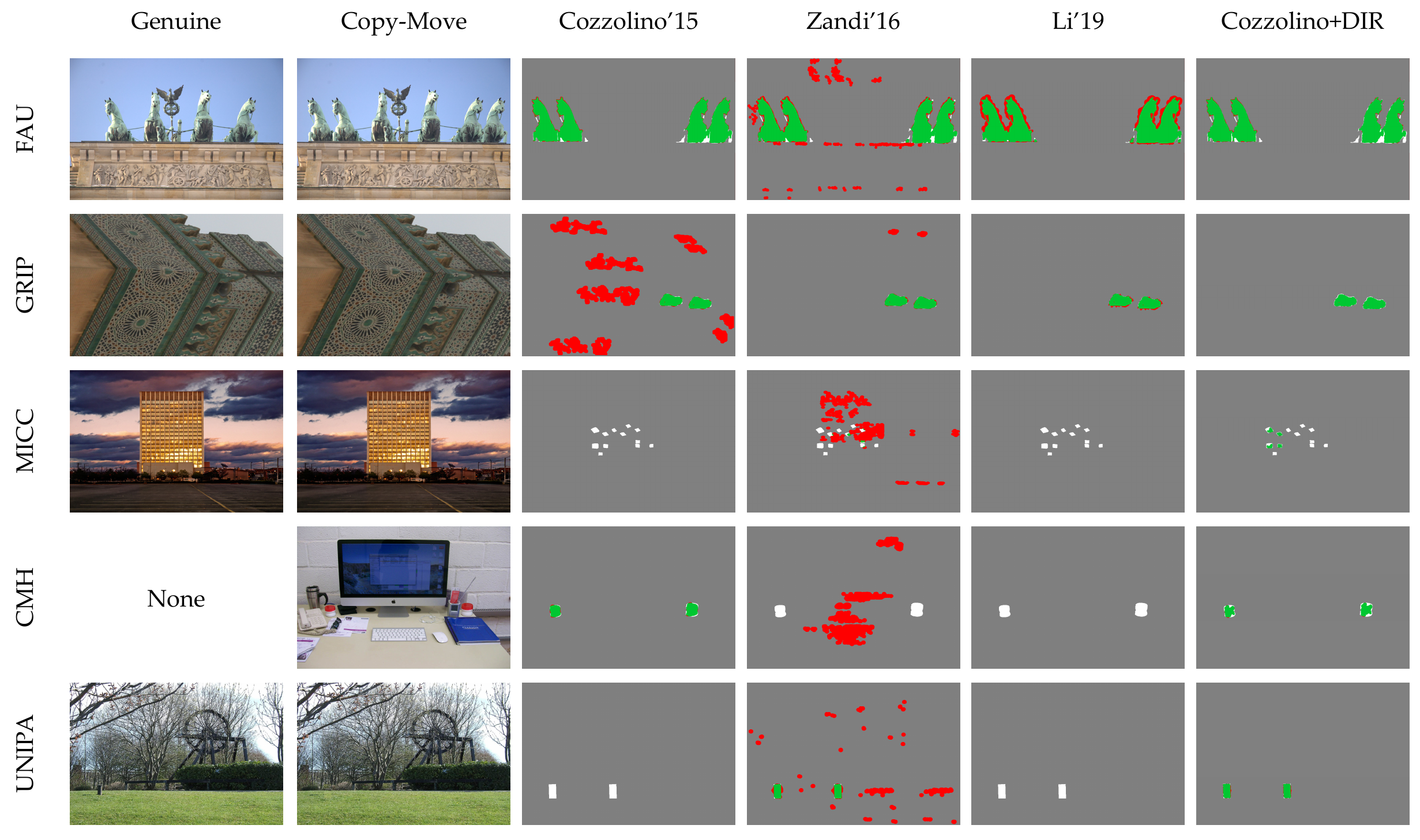}
	\centering
	\caption{Some samples of the copy-move detection by different forensic methods on the copy-move forensic benchmarks.}
\end{figure*}

\begin{figure*}[!t]
	\centering
	\includegraphics[scale=0.59]{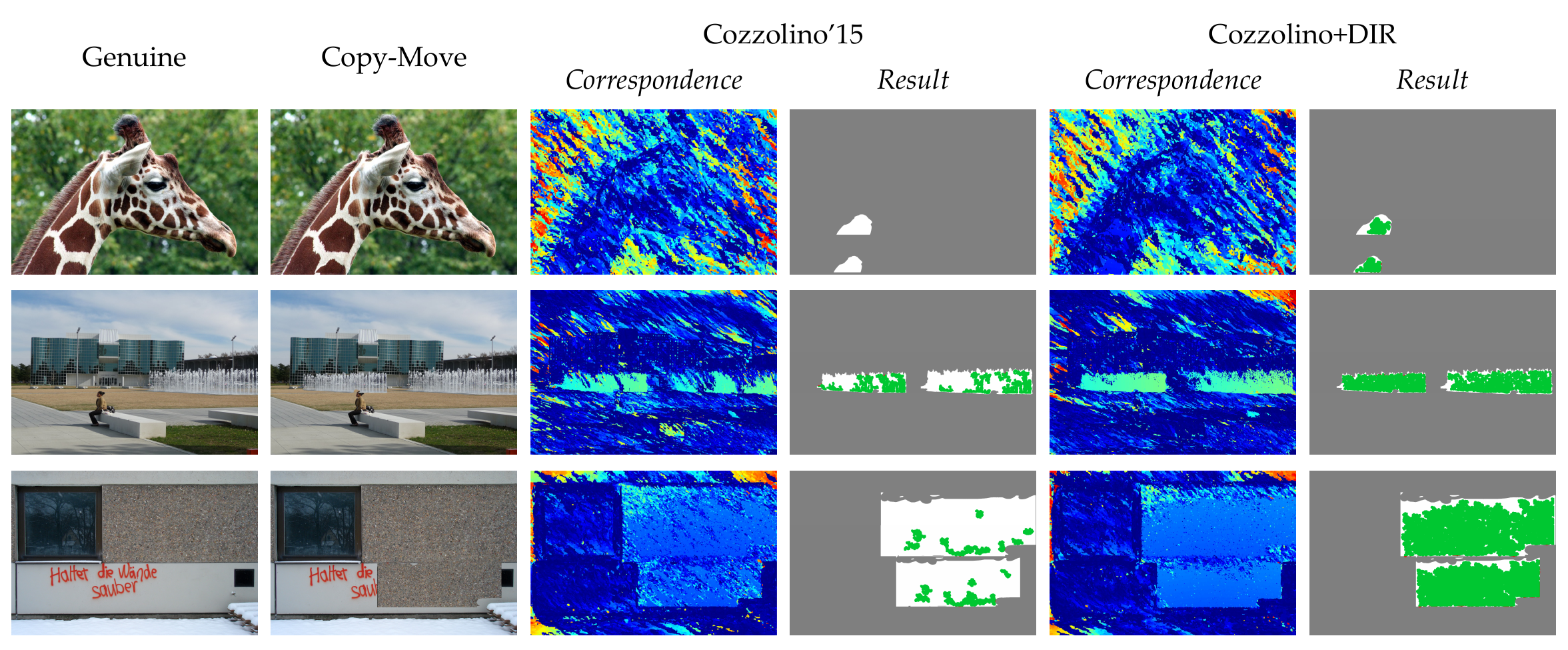}
	\centering
	\caption{Some samples of the copy-scale-move detection by different forensic methods.}
\end{figure*}

\emph{Robustness and Invariance}. Fig. 6 shows some samples of the offset length map exported from PatchMatch, while the repeatability scores for all above comparison methods are given in Table 4. Regarding geometric variations, the DAISY and DASC are difficult to maintain a stable matching, especially for the flipping and large-angle rotation. The DAISY exhibits a good repeatability for scaling, while the DASC does not have such a property. As for signal corruptions, the overall repeatability of DASC is better than DAISY, which is consistent with the design goal of DASC. In general, compared with such well-designed descriptors for the matching task, our DIR is typically more robust to geometric variations and provides comparable repeatability against signal corruptions.

\begin{table}[bp]
	\centering
	\caption{Precision, Recall, and F1 Scores (\%) for Different Methods on Various Copy-Move Forensic Benchmarks.}
	\begin{tabular}{cccccc}
		\toprule
		\multicolumn{2}{c}{Method} & \multicolumn{1}{c}{\tabincell{c}{Cozzolino’15 \\ \cite{ref44}}} & \multicolumn{1}{c}{\tabincell{c}{Zandi’16 \\ \cite{ref48}}} & \multicolumn{1}{c}{\tabincell{c}{Li’19 \\ \cite{ref49}}} & \multicolumn{1}{c}{\tabincell{c}{Cozzolino\\+DIR}} \\
		\midrule
		\multicolumn{1}{c}{\multirow{5}[1]{*}{\begin{sideways}Precision\end{sideways}}} & FAU   & 93.08 & 79.59 & 89.44 & 96.75 \\
		& GRIP  & 93.01 & 81.9  & 92.7  & 96.45 \\
		& MICC & 92.02 & 66.06 & 88.77 & 94.3 \\
		& CMH   & 82.98 & 54.96 & 85.26 & 88.62 \\
		& UNIPA & 85.37 & 73.07 & 86.18 & 96.5 \\
		\midrule
		\multicolumn{1}{c}{\multirow{5}[1]{*}{\begin{sideways}Recall\end{sideways}}} & FAU   & 91.94 & 95.09 & 88.94 & 92.44 \\
		& GRIP  & 96.41 & 98.64 & 97.43 & 95.89 \\
		& MICC & 89.07 & 74.98 & 86.53 & 88.76 \\
		& CMH   & 78.99 & 65.27 & 71.98 & 75.91 \\
		& UNIPA & 91.76 & 99.05 & 93.95 & 95.03 \\
		\midrule
		\multicolumn{1}{c}{\multirow{5}[1]{*}{\begin{sideways}F1\end{sideways}}} & FAU   & 91.89 & 83.39 & 88.63 & 93.62 \\
		& GRIP  & 93.85 & 86.67 & 94.75 & 95.74 \\
		& MICC & 89.22 & 66.41 & 86.73 & 89.82 \\
		& CMH   & 80.12 & 58.25 & 76.35 & 79.5 \\
		& UNIPA & 88.31 & 81.01 & 89.3  & 95.66 \\
		\bottomrule
	\end{tabular}%
\end{table}%

\emph{Efficiency}. Turning to efficiency, the feature extraction time for DAISY, DASC, and DIR is $ \sim $ 6 seconds, $ \sim $ 106 seconds, and $ \sim $ 6 seconds, respectively; the feature matching time for DAISY, DASC, and DIR is $ \sim $ 79 seconds, $ \sim $ 77 seconds, and $ \sim $ 20 seconds, respectively. Owing to the proposed constant-time implementation, the feature extraction of DIR is quite fast and comparable to DAISY, which is well known for its efficiency. In addition, the feature vector of DIR is more compact, due to the orthogonality of basis functions, allowing a significant time saving in the matching process.

\subsection{Passive Forensics: Copy-Move Forgery Detection}

Copy-move is one of the most basic operations in image forgery. It involves copying and pasting specific patches from and to an image. Typical detection algorithm extracts sparse or dense features from the image and reveals potential copy-move regions by matching such features. Obviously, the quality of the extracted features has a significant impact on the performance. In general, sparse methods are more efficient and geometrically invariant but exhibit lower accuracy; conversely, dense methods are more accurate but relatively less efficient and geometrically invariant. This phenomenon is consistent with our analysis in Section 1.1.

\begin{table*}[t]
	\centering
	\caption{Precision, Recall, and F1 Scores (\%) for Different Methods on the FAU Copy-Move Forensic Benchmark.} 
	\begin{threeparttable}
		\begin{tabular}{cccccccccc}
			\toprule
			Method & Ryu’13 \cite{ref50} & Li’13 \cite{ref51} & Silva’15 \cite{ref46} & Emam’16 \cite{ref52} & Pun’18 \cite{ref53} & Bi’18 \cite{ref54} & Wu’18 \cite{ref55} & Zhong’20 \cite{ref56} & Cozzolino+DIR \\
			\midrule 
			Precision & 95.02 & 58.05 & 88.02 & - & 91.07 & 90.55 & 44.59 & 75.61 & 96.75 \\ 
			Recall & 88.15 & 92.26 & 89.72 & - & 90.21 & 91.66 & 31.60 & 74.13 & 92.44 \\ 
			F1 & 91.50 & 71.21 & 88.95 & 84.91 & 90.33 & 91.07 & 37.11 & 74.82 & 93.62 \\
			\bottomrule 
		\end{tabular}
		\begin{tablenotes}
			\footnotesize
			\item \emph{The results for comparison methods in this table are cited directly from \cite{ref56}.}
		\end{tablenotes}
	\end{threeparttable}
\end{table*}

\begin{table}[t]
	\caption{Precision, Recall, F1 Scores (\%), and Matching Performance (Number of Matches per Image) Gain Rate in Copy-Scale-Move Robustness Experiment.}
	\centering
	\begin{tabular}{cccc}
		\toprule
		Method & \tabincell{c}{Cozzolino’15 \\ \cite{ref44}} & Cozzolino+DIR & Gain rate \\ 
		\midrule
		Precision & 84.47 & 90.86 & 7.56 \\ 
		Recall & 46.06 & 63.32 & 37.47 \\ 
		F1 & 56.15 & 71.18 & 26.77 \\ 
		\#matches/image & 144519.75 & 270730.23 & 87.33 \\ 
		\bottomrule
	\end{tabular}
\end{table}

\begin{table*}[bp]
	\caption{Precision, Recall, and F1 Scores (\%) for Different Methods in Copy-Scale-Move Robustness Experiment.}
	\centering
	\begin{threeparttable}
		\begin{tabular}{ccccccccc}
			\toprule
			Method & Ryu’13 \cite{ref50} & Li’13 \cite{ref51} & Emam’16 \cite{ref52} & Pun’18 \cite{ref53} & Bi’18 \cite{ref54} & Wu’18 \cite{ref55} & Zhong’20 \cite{ref56} &  Cozzolino+DIR \\ \midrule
			Precision & $<$ 25 & $<$ 25  & - & 41.48 & 62.09 & 34.84 & 68.05 & 90.86 \\
			Recall & $<$ 20 & 24.95 & - & 39.69 & $<$ 20 & 20.12 & 64.59 & 63.32 \\ 
			F1 & $<$ 20 & 20.97 & $<$ 20 & 40.51 & 22.91 & 25.22 & 64.67 & 71.18 \\ 
			\bottomrule
		\end{tabular}
		\begin{tablenotes}
			\footnotesize
			\item \emph{The results for comparison methods in this table are cited directly from \cite{ref56}.}
		\end{tablenotes}
	\end{threeparttable}
\end{table*}

\emph{Algorithm Design}. A representative work of the dense approach is proposed by Cozzolino et al. \cite{ref44}, relying on rotation-invariant orthogonal moments. Theoretically, its feature extraction module can be considered as a special case of DIR, with a fixed scale. As can be expected, in practice, this method is stable under rotation, flipping and signal corruptions, but sensitive to scale changes. We hence introduce the DIR framework into this algorithm mainly for improving its scaling robustness. Specifically, the original PCT feature extraction module is directly replaced by the DIR extension of PCT. For implementation details of DIR, we chose a set of $(n,m)$ with a maximum order of 3, generating 10-dimensional features for each position $(u,v,w)$ in the scale space. Here, we consider a dense sampling of $(u,v)$ over the image grid, and the scale $w$ increases from 8 to 32 with 10 samples. For a compact representation with scaling tolerance, the feature vectors are average-pooled together over the scales. As a common trick, the low-resolution image will be upsampled (long edge = 2000 pixels) to suppress the parameter sensitivity. Note that we exclude the border pixels between forgery and background in the score computation \cite{ref23, ref44}.

\emph{Copy-Move Benchmark}. Firstly, we perform a quantitative comparison on five copy-move forensic benchmarks: FAU \cite{ref23}, GRIP \cite{ref44}, MICC \cite{ref45}, CMH \cite{ref46}, and UNIPA \cite{ref47}. Here, FAU, GRIP, and UNIPA contain only rigid copy-move manipulation; while MICC and CMH are with further attacks (e.g., scaling and rotation) for a convincing visual effect. The experiment involves the basic Cozzolino’15 \cite{ref44} and its DIR version, as well as state-of-the-art sparse algorithms: Silva’15 \cite{ref46}, Zandi’16 \cite{ref48} and Li’19 \cite{ref49}; state-of-the-art dense algorithms: Ryu’13 \cite{ref50}, Li’13 \cite{ref51}, Emam’16 \cite{ref52}, Pun’18 \cite{ref53}, Bi’18 \cite{ref54}, Wu’18 \cite{ref55}, Zhong’20 \cite{ref56}. Here, Wu’18 and Zhong’20 are based on deep neural networks. Note that the feature extraction modules in Ryu’13, Li’13, Emam’16, Pun’18, and Bi’18 can also be considered as special cases of DIR.

Fig. 7 shows some samples of copy-move detection on benchmarks. The precision, recall, and F1 scores for popular open-source algorithms \cite{ref44, ref48, ref49} over all above benchmarks are given in Table 5. For the rest of the comparison methods, the scores are summarized in Table 6, only on the FAU due to the lack of the code. It can be observed that, in general, the dense approach, especially Cozzolino’15 and its DIR extension, provides higher detection accuracy than the sparse approach. As far as the sparse approach is concerned, the Li’19 exhibits the performance advantage over the Zandi’16, which can be mainly attributed to the more dense keypoints. Such phenomenon confirms that the coverage is a vital factor in the image representation for forensics. With the introduction of DIR, the accuracy of Cozzolino’15 on most benchmarks is further improved (basically the same on CMH), meaning that DIR has practical usefulness rather than just as a mathematical extension. Such accuracy gain is largely due to the multi-scale framework in our DIR, which allows features to be more informative also with scaling tolerance.

\begin{figure*}[!t]
	\centering
	\subfigure{\includegraphics[scale=0.5]{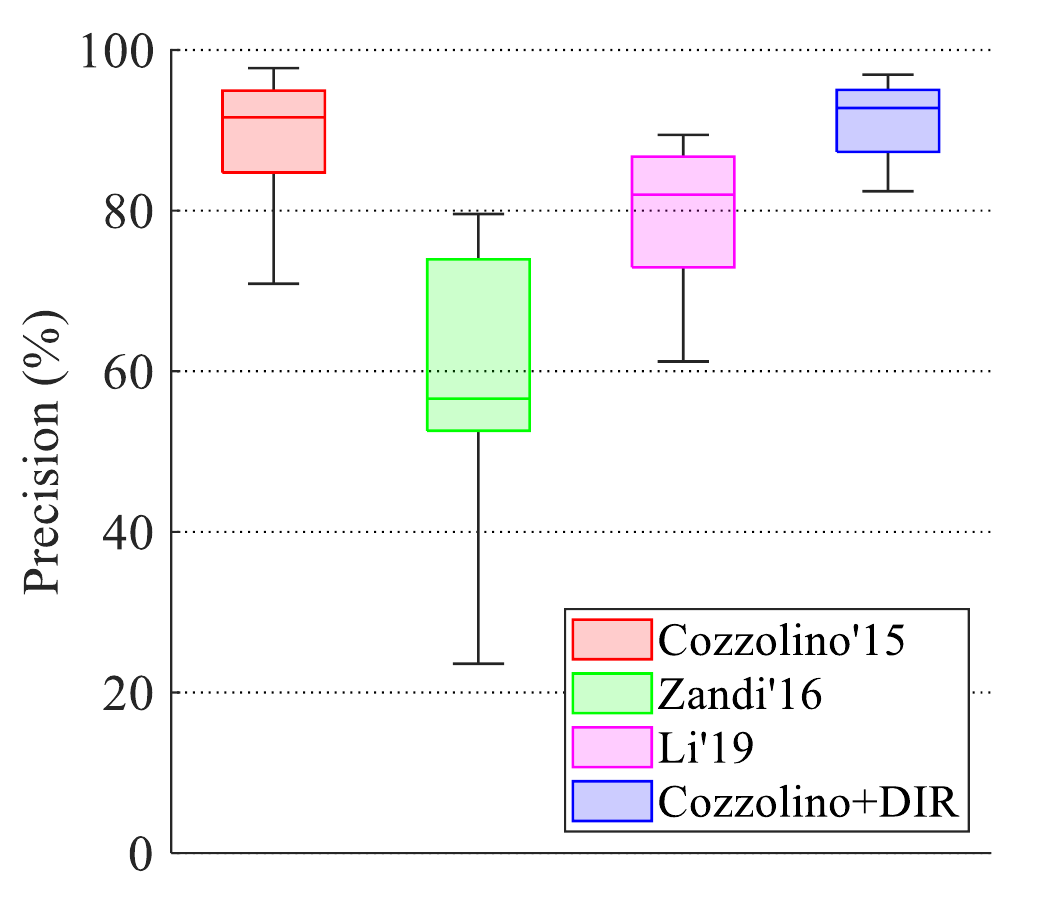}}
	\subfigure{\includegraphics[scale=0.5]{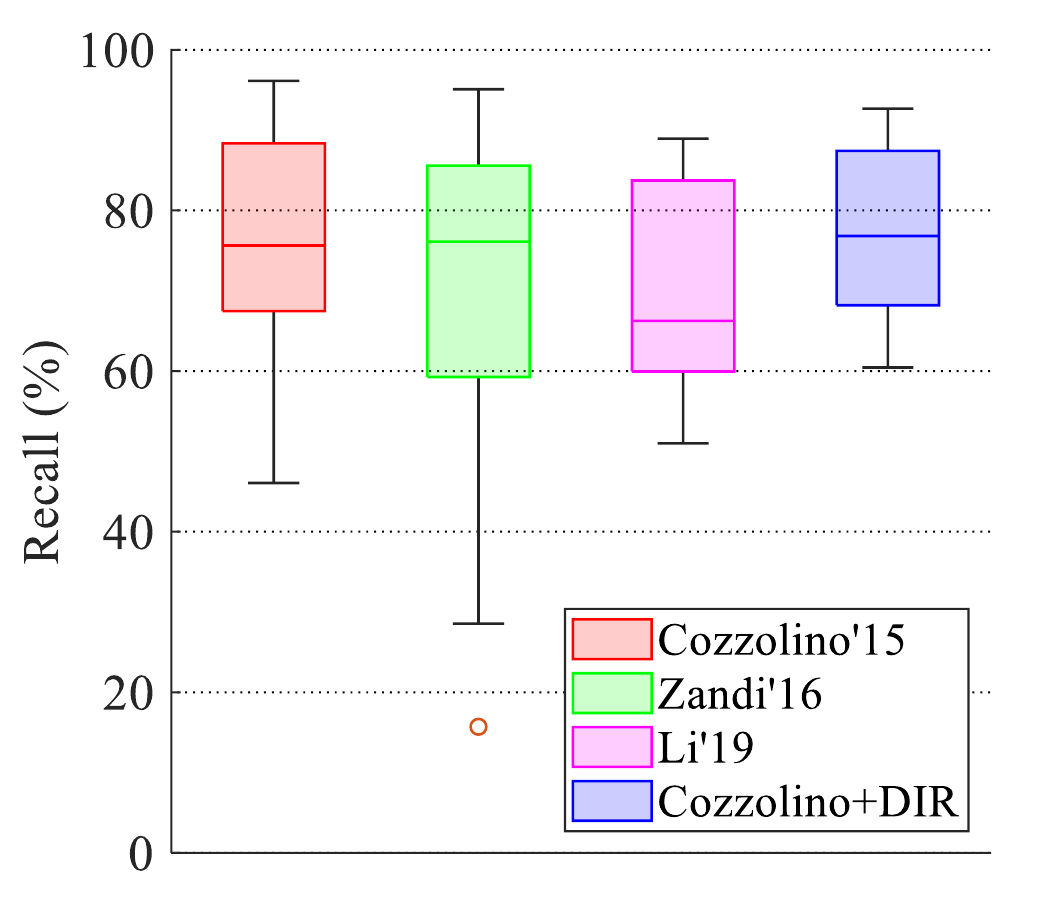}}
	\subfigure{\includegraphics[scale=0.5]{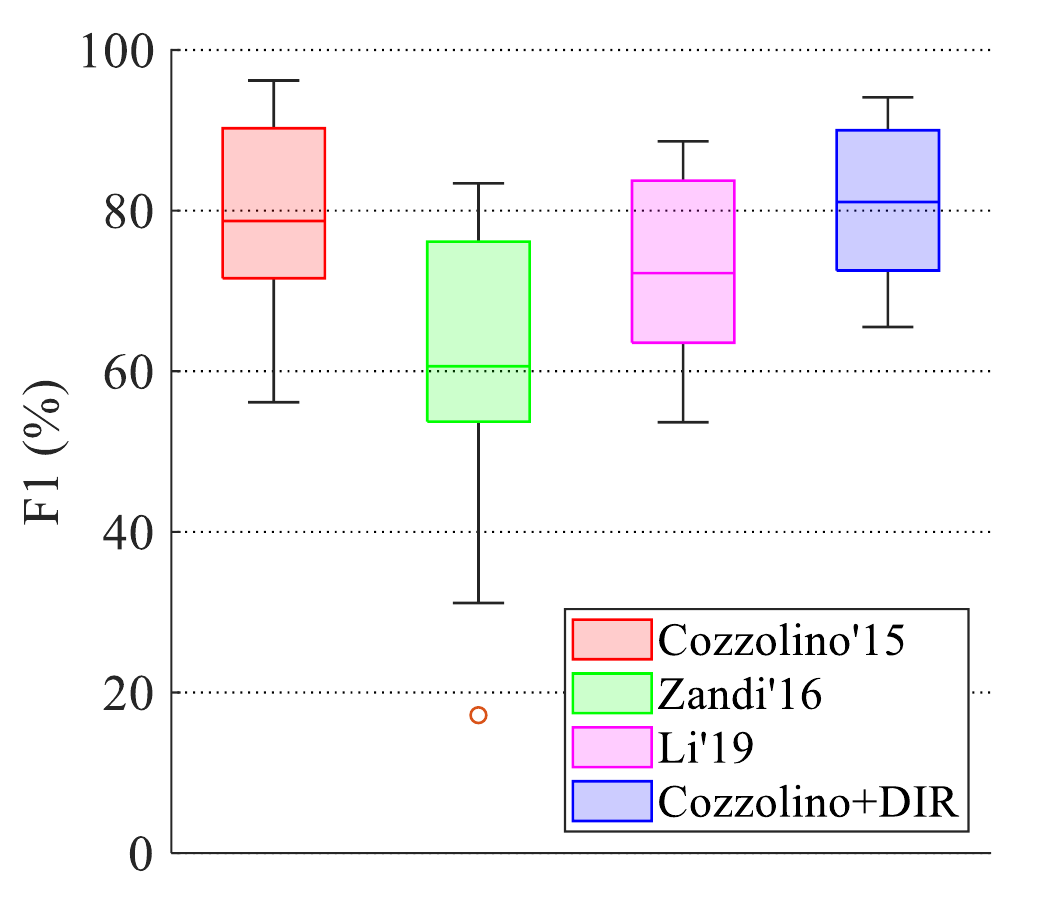}}
	\centering
	\caption{Precision, recall, and F1 box-plots by different copy-move forgery detection methods in comprehensive robustness experiment.}
\end{figure*}

\emph{Copy-Scale-Move Robustness}. Secondly, we evaluate the scaling tolerance of the basic Cozzolino’15 and its DIR version. The experiment is designed on FAU benchmark with copy-scale-move manipulation, where the factor is set to 0.8. The experiment also covers some state-of-the-art dense algorithms as comparison baselines: Ryu’13 \cite{ref50}, Li’13 \cite{ref51}, Emam’16 \cite{ref52}, Pun’18 \cite{ref53}, Bi’18 \cite{ref54}, Wu’18 \cite{ref55}, and Zhong’20 \cite{ref56}.

Fig. 8 shows some samples of the correspondence map and detection result for copy-scale-move. The performance statistics are summarized in Tables 7 and 8. In the scenario with scale variations, the accuracy gap between Cozzolino’15 and its DIR version widens significantly. Also, strong performance degradation is generally observed for other dense algorithms. The introduction of DIR improves the recall ($\sim$ 37\% gain) and the precision ($\sim$ 8\% gain), thus exhibiting a higher F1 ($ \sim $ 27\% gain). The more noticeable fact is that the average number of matches per image has nearly $\times 2$ in the DIR version. In addition, the proposed algorithm also demonstrates benefits compared to the similar dense strategies.

\emph{Comprehensive Robustness}. Finally, we evaluate the robustness under comprehensive attacks, involving six signal corruptions of whole images and four geometric transformations of manipulated regions, on the FAU benchmark. The comparison methods include  Cozzolino’15 \cite{ref44},  Zandi’16 \cite{ref48}, and Li’19 \cite{ref49}. To reveal the distribution properties of the forensic scores (especially the average and deviation nature), the corresponding box-plots are presented in Fig. 9. Note that we provide the detailed scores for each attack in Appendix F.

It is clear from Fig. 9 that, compared with Cozzolino’15 in precision, recall, and F1 metrics, the proposed DIR version exhibits lower score deviations while maintaining higher or similar average scores. As for Zandi’16 and Li’19, greater performance gain is achieved by our method from both average and deviation perspectives. These common phenomena confirm that, with the introduction of our DIR, the stability of the copy-move forensic algorithm is significantly improved (especially for scaling) while not compromising the average-level of accuracy.

Such experimental evidence supports our theoretical expectation on geometric transformation robustness, verifying the usefulness of DIR for passive forensic scenarios. We would like to make a comment that the above application of DIR is naive; more careful design for scaling invariance can be achieved by incorporating the dense scale selection.

\begin{figure*}[!t]
	\centering
	\includegraphics[scale=0.8]{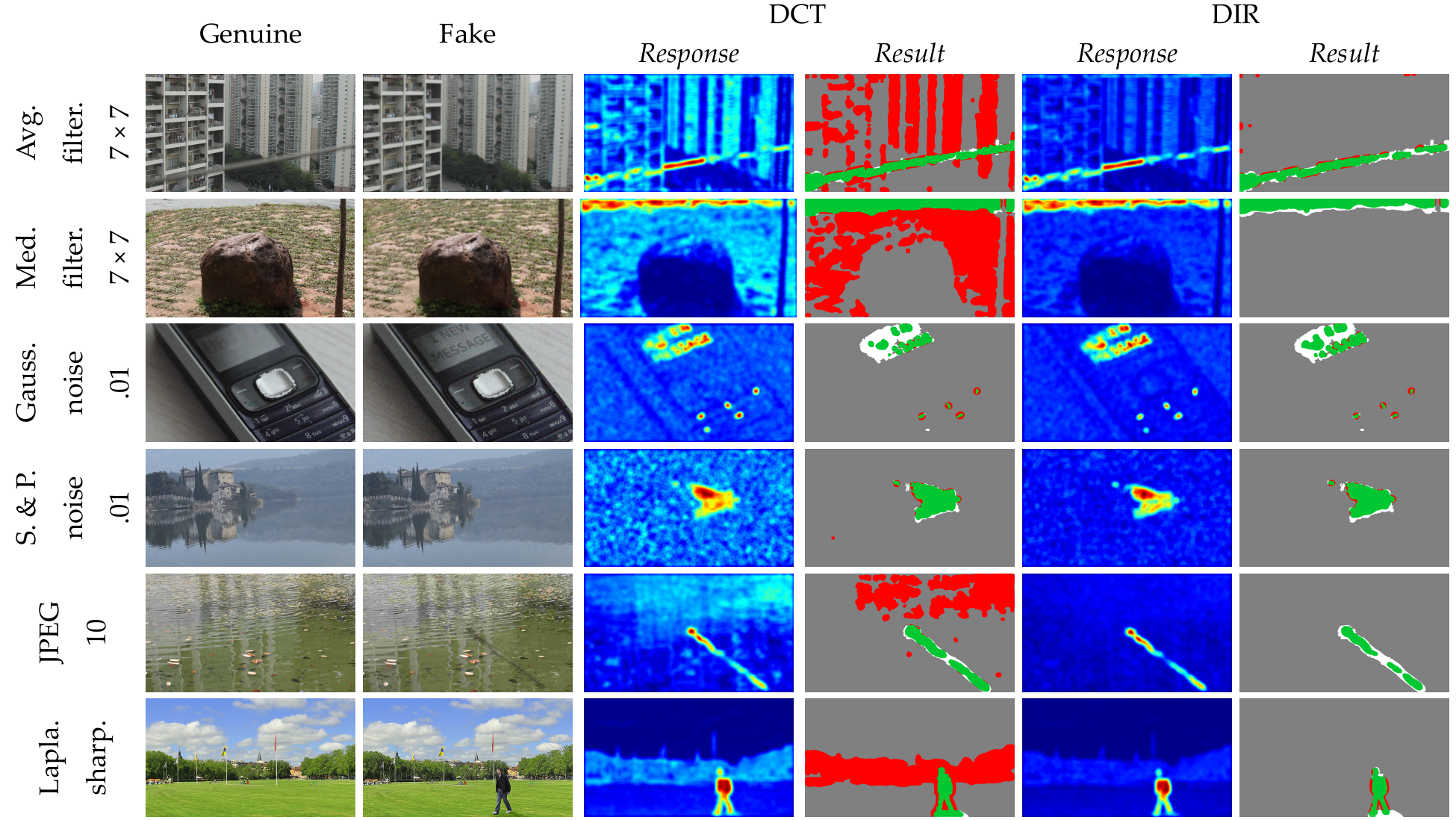}
	\centering
	\caption{Some samples of the content authentication by different perceptual hashing methods.}
\end{figure*}

\subsection{Active Forensics: Perceptual Hashing}

Perceptual hashing is a well-known active forensic framework for multimedia content authentication. The main idea relies on deriving a compact hash sequence as digital abstract of the image content. With this idea in mind, the image representation should be robust to content-preserving operations and discriminative to visually distinct contents; as for efficiency, the hash sequence is expected to be compact enough. Note that such requirements are also consistent with the discussion in Section 1.1.

\emph{Algorithm Design}. Currently, the state-of-the-art perceptual hashing algorithms generally follow a general framework that combines sparse and dense features, where sparse features (e.g., SIFT) for geometric correction and dense features (e.g., dense DCT) for forgery localization \cite{ref59, ref62}. Despite their popularity, the DCT-like dense features still suffer from flaws in both representation capability and hash compactness. The forensic results are quite unstable under certain signal corruptions, such as blurring and compression. Motivated by this, we replace such dense DCT by our DIR for improving its signal corruption robustness while shortening the hash sequence. Here, we consider an 8-pixel sampling interval for $(u,v)$ over the image grid, which is a common setting for compactness. The order constraint is set as $K = 3$ in (11) with $\infty $-norm, resulting in 16-dimensional features for each position. The scale $w$ only increases from 8 to 12 with 3 samples due to the presence of geometric correction, and such feature vectors are then average-pooled together over the scales. In the implementation, the threshold for hash distance comparison is important. For a fair experiment, this threshold is determined by the adaptive Otsu's method \cite{ref57} for all comparison methods.

\begin{table*}
	\centering
	\caption{Precision, Recall, and F1 Scores (\%) for Different Perceptual Hashing Algorithms in Hash Robustness Experiment.}
	\begin{tabular}{ccccccc}
		\toprule
		\multirow{2}[4]{*}{Method} & \multicolumn{3}{c}{DCT \cite{ref59, ref62}} & \multicolumn{3}{c}{DIR} \\
		\cmidrule{2-7}    \multicolumn{1}{c}{} & \multicolumn{1}{c}{Precision} & \multicolumn{1}{c}{Recall} & \multicolumn{1}{c}{F1} & \multicolumn{1}{c}{Precision} & \multicolumn{1}{c}{Recall} & \multicolumn{1}{c}{F1} \\
		\midrule
		Clean & 87.24 & 77.7  & 79.85 & 89.03 & 71.97 & 76.95 \\
		\midrule
		Gaussian noise 0.01 & 87.02 & 74.81 & 77.8  & 88.63 & 70.13 & 75.46 \\
		Gaussian noise 0.02 & 86.74 & 74.2  & 77.21 & 87.8  & 70.05 & 74.68 \\
		Salt and pepper noise 0.01 & 86.38 & 77.83 & 79.52 & 88.61 & 71.88 & 76.68 \\
		Salt and pepper noise 0.02 & 85.73 & 77.66 & 78.97 & 88.29 & 71.65 & 76.35 \\
		Average filtering 7 × 7 & 55.43 & 75.33 & 53.43 & 78.8  & 68.09 & 68.06 \\
		Average filtering 14 × 14 & 37.87 & 74.02 & 38.45 & 52.89 & 65.55 & 50.18 \\
		Gaussian filtering 7 × 7 & 59.49 & 75.48 & 56.72 & 80.43 & 68.73 & 69.41 \\
		Gaussian filtering 14 × 14 & 49.86 & 74.97 & 49.03 & 71.47 & 68.17 & 63.35 \\
		Median filtering 7 × 7 & 60.61 & 74.91 & 57.84 & 82.43 & 68.51 & 70.17 \\
		Median filtering 14 × 14 & 44.72 & 69.49 & 43.13 & 66.83 & 61.51 & 56.72 \\
		JPEG compression 10 & 84.22 & 76.08 & 76.28 & 88.61 & 71.02 & 76.15 \\
		JPEG compression 5 & 77.12 & 75.57 & 69.86 & 87.55 & 70.88 & 75.29 \\
		Laplacian sharpening & 55.36 & 80.17 & 54.55 & 78.76 & 72.79 & 69.64 \\
		\midrule
		\emph{Average} $\uparrow$ & 68.41 & 75.59 & 63.76 & 80.72 & 69.35 & 69.94 \\
		\emph{Standard deviation} $\downarrow$ & 17.53 & 2.36  & 14.32 & 10.26 & 2.89  & 7.88 \\
		\bottomrule
	\end{tabular}
\end{table*}

\emph{Hash Robustness}. A quantitative comparison is performed on forensic benchmark RTD \cite{ref58}, which involves not only the inpainting, splicing, and copy-move manipulations, but also other subtle changes by hand. Due to the inherent nature of perceptual hashing, the forensics on such clean images is effortless. Therefore, we mainly consider the more challenging tasks under global signal corruptions. The comparison methods include the basic SIFT-DCT framework and our SIFT-DIR version. Note that such basic framework is involved with several state-of-the-art perceptual hashing algorithms, e.g., Zhang’20 \cite{ref59}, Wang’15 \cite{ref62}, Hao’21 \cite{ref60}, and Biswas’20 \cite{ref61}.

Fig. 10 shows some samples of the response map and detection result, while the precision, recall, and F1 scores are summarized in Table 9. One can note that, even in the forensic scenarios employing geometric correction, the DIR still exhibits certain advantages over common descriptors. As shown in the response map, under some operations, it is challenging for DCT to be robust while maintaining discriminability. The DCT typically exhibits false positives in regions with content-preserving operations. In general, compared with such state-of-the-art framework, our DIR version provides comparable detection accuracy for clean and noisy images, while generally being more robust to blurring, compression, and sharpening. The average and standard deviation of the scores also demonstrate the advantage of DIR, especially in terms of precision (w.r.t. false positives). The DIR exhibits better overall localization accuracy (for average) and is more stable under different attacks (for standard deviation).

\emph{Hash Compactness}. Turning to compactness, the dimensions of DCT and DIR features on each position are 32 and 16, respectively. This means a nearly 50\% saving in terms of storage and transmission costs.

Such experimental evidence supports our theoretical expectation on signal perturbation robustness and representation compactness, verifying the usefulness of DIR for active forensic scenarios.

\section{Conclusion}
The main goal of this paper is to provide a principled design of image representation for forensic tasks. We name this complete pipeline “Dense Invariant Representation”, meaning a local representation with covariance for position, orientation and scale variations.

The key ingredients of our work are as follows. i) At the theoretical level, the global definition of classical orthogonal moments is extended to the local with scale space. With such generic definition, the deformation stability backed with “invariance-equivariance-covariance” framework is explicitly analyzed (Section 3). ii) At the implementation level, the fine-tuned discrete computation strategy is designed, which is characterized by low integral/numerical error, constant complexity, and generality for arbitrary basis functions (Section 4). iii) At the application level, the above ideas are fully validated in two vision tasks (dense pattern detection and matching) and two forensic tasks (copy-move forgery detection and perceptual hashing). In general, our method gives state-of-the-art accuracy and efficiency performance, proving its promise in small-scale robust vision problems (Section 6).

The limitations for current approach includes scaling invariance construction and space complexity. We note that the common pooling operations only provide a limited tolerance for scaling. A more elegant construction is in our plan, relying on a suitable fusion of DIR and some recent advances of dense scale selection. Additionally, our fast implementation based on the convolution theorem may increase the space complexity. As a long-term research path, such problem is expected to be mitigated by hardware-oriented optimization and more careful adaptive time-space trade-offs.

%

%
%
%
%
\ifCLASSOPTIONcompsoc
  \section*{Acknowledgments}
\else
  \section*{Acknowledgment}
\fi

This work was supported in part by the National Key R\&D Program of China under Grant 2019YFB1406500, in part by the Postgraduate Research \& Practice Innovation Program of Jiangsu Province under Grant KYCX22\_0383, in part by the National Natural Science Foundation of China under Grants 62072237, 61971476, and U2001202, in part by Guangxi Key Laboratory of Trusted Software under Grant KX202027, in part by Basic Research Program of Jiangsu Province under Grant BK20201290, in part by Macau Science and Technology Development Fund under Grants SKLIOTSC-2021-2023 and 0072/2020/AMJ, and in part by Research Committee at University of Macau under Grant MYRG2020-00101-FST.

\appendices

\section{Common Definitions of Radial Basis Functions in (4)}
In Table A1, we include some common definitions of radial basis functions for (4), with orthogonality and completeness \cite{ref32}. The main discussion in this paper is generic to all such definitions. Also, we implement all the DIR extensions for the methods listed in Table A1. The code is available online at \texttt{https://github.com/ShurenQi/DIR}.

\section{Proof of (8): Equivariance to Translation}
By plugging $f_T$ into (7), one can check that image translation operation only affects the representation parameters $(u,v)$, as follows:
\begin{equation}\nonumber
	\begin{split}
		&\left< {f_T}(x,y),V_{nm}^{uvw}(x,y) \right>\\
		&= \iint\limits_D{R_n^*(\frac{1}{w}\sqrt {{{(x - u)}^2} + {{(y - v)}^2}} )A_m^*(\arctan (\frac{{y - v}}{{x - u}}))}\\
		&\times {f(x + \Delta x,y + \Delta y)dxdy} \\
		&= \iint\limits_D{R_n^*(\frac{1}{w}\sqrt {{{(x - \Delta x - u)}^2} + {{(y - \Delta y - v)}^2}} )}\\
		&\times {A_m^*(\arctan (\frac{{y - \Delta y - v}}{{x - \Delta x - u}}))f(x,y)dxdy}\\
		&=\iint\limits_D{R_n^*(\frac{1}{w}\sqrt {{{(x - (u + \Delta x))}^2} + {{(y - (v + \Delta y))}^2}} )}\\
		&\times	{A_m^*(\arctan (\frac{{y - (v + \Delta y)}}{{x - (u + \Delta x)}}))f(x,y)dxdy}\\
		&= \left< f(x,y),V_{nm}^{(u + \Delta x)(v + \Delta y)w}(x,y) \right>,
	\end{split}
\end{equation}
where the same offset $(\Delta x,\Delta y)$ also appears in representation $\left< f(x,y),V_{nm}^{(u + \Delta x)(v + \Delta y)w}(x,y) \right>$, implying the equivariance w.r.t. translation.

\section{Proof of (9): Covariance and Invariance to Rotation }
By plugging ${f_R}$ into (7) with  $(u,v) = (0,0)$ and writing down in polar form, one can check that image rotation operation only affects the phase of representation, as follows:
\begin{equation}\nonumber
	\begin{split}
		&\left< {f_R}(r,\theta),V_{nm}^{uvw}(r',\theta ') \right> \\
		&= \iint\limits_D{R_n^*(r')A_m^*(\theta ')f(r,\theta  + \phi )r'dr'd\theta '}\\
		&= \frac{1}{{{w^2}}}\int\limits_0^{2\pi } {\int\limits_0^w {R_n^*(\frac{r}{w})A_m^*(\theta )f(r,\theta  + \phi )rdrd\theta } } \\
		&= \frac{1}{{{w^2}}}\int\limits_0^{2\pi } {\int\limits_0^w {R_n^*(\frac{r}{w})A_m^*(\theta  - \phi )f(r,\theta )rdrd\theta } } \\
		&= \frac{1}{{{w^2}}}\int\limits_0^{2\pi } {\int\limits_0^w {R_n^*(\frac{r}{w})A_m^*(\theta )A_m^*( - \phi )f(r,\theta )rdrd\theta } } \\
		&= \frac{1}{{{w^2}}} \left< f(r,\theta ),V_{nm}^{uvw}(r',\theta ') \right> {w^2}\;A_m^*( - \phi ) \\
		&= \left< f(r,\theta ),V_{nm}^{uvw}(r',\theta ') \right> A_m^*( - \phi ),
	\end{split}
\end{equation}
where $A_m^*(\theta  - \phi ) = A_m^*(\theta )A_m^*( - \phi )$ is true as ${A_m}(\theta ) = \exp (\bm{j}m\theta )$; the same angle $\phi$ also appears in phase of the representation $\left< f(r,\theta ),V_{nm}^{uvw}(r',\theta ') \right> A_m^*( - \phi )$, implying the covariance w.r.t. rotation. Therefore, the magnitude-only strategy is able to derive the rotation-invariant features: $| \left< {f_R}(r,\theta),V_{nm}^{uvw}(r',\theta ') \right>|=| \left< f(r,\theta ),V_{nm}^{uvw}(r',\theta ') \right>|$.

\section{Proof of (10): Covariance to Scaling }
By plugging ${f_S}$ into (7) with $(u,v) = (0,0)$, one can check that image scaling operation only affects the representation parameters $w$, as follows:
\begin{equation}\nonumber
	\begin{split}
		&\left< {f_S}(x,y),V_{nm}^{uvw}(x,y) \right>\\
		&= \iint\limits_{{x^2} + {y^2} \le {w^2}}{R_n^*(\frac{1}{w}\sqrt {{x^2} + {y^2}} )A_m^*(\arctan (\frac{y}{x}))f(sx,sy)dxdy}\\
		&= \iint\limits_{{x^2} + {y^2} \le {w^2}{s^2}}{R_n^*(\frac{1}{w}\sqrt {{{(\frac{x}{s})}^2} + {{(\frac{y}{s})}^2}} )A_m^*(\arctan (\frac{y}{x}))}\\
		&\times {f(x,y)d\frac{x}{s}d\frac{y}{s}}\\
		&= \frac{1}{{{s^2}}}\iint\limits_{{x^2} + {y^2} \le {w^2}{s^2}}{R_n^*(\frac{1}{{ws}}\sqrt {{x^2} + {y^2}} )A_m^*(\arctan (\frac{y}{x}))}\\
		&\times f(x,y)dxdy\\
		&= \frac{1}{{{s^2}}}\left< f(x,y),V_{nm}^{uv(ws)}(x,y) \right>{s^2}\\
		&= \left< f(x,y),V_{nm}^{uv(ws)}(x,y) \right>,
	\end{split}
\end{equation}
where it should be reminded that the area for integration changes with the factor of ${s^2}$; the same factor $s$ also appears in the representation $\left< f(x,y),V_{nm}^{uv(ws)}(x,y) \right>$, implying the covariance w.r.t. scaling.

\section{An Analysis of the Robustness to Signal Corruption}

In fact, our DIR exhibits a certain robustness to some signal corruptions, e.g., additive noise and low-pass filtering/blurring. This is mainly inherited from the classical orthogonal moment theory, i.e., the low-order moments are less affected by the signal corruptions that act on the high-frequency components of the image. Here, we give a formal analysis of this property from a frequency-domain perspective.

Without loss of generality, the signal corruptions of the modern imaging process can be modeled as:
\begin{equation}\nonumber
	f = p \otimes s + n,
\end{equation}
where $f$ is the captured image, $s$ is the ideal scene (i.e., ground-truth image), $p$ denotes the Point-Spread Function (PSF) w.r.t. image blurring, $n$ denotes the additive noise, and $ \otimes $ denotes the spatial convolution operation. Here, we assume that PSF $p$ is a common low-pass filter, e.g., a Gaussian kernel function, and $n$ is random white noise with limited intensity.

Considering the DIR of $f$ with dense sampling of $(u,v)$ over the discrete grid, it can be written as follows:
\begin{equation}\nonumber
	\left< f,V_{nm}^{uvw} \right> = f(i,j) \otimes {(H_{nm}^w(i,j))^T}.
\end{equation}

By plugging $f = p \otimes s + n$ into this equation, we can derive an expanded version as follows:
\begin{equation}\nonumber
	\begin{split}
		\left< f,V_{nm}^{uvw} \right> &= f \otimes {(H_{nm}^w)^T}\\
		&= (p \otimes s + n) \otimes {(H_{nm}^w)^T}\\
		&= p \otimes s \otimes {(H_{nm}^w)^T} + n \otimes {(H_{nm}^w)^T}.\\
	\end{split}
\end{equation}

Let us analyze this result from a frequency-domain perspective:
\begin{equation}\nonumber
	\begin{split}
		& \mathcal{F}(p \otimes s \otimes {(H_{nm}^w)^T} + n \otimes {(H_{nm}^w)^T})\\
		& = \mathcal{F}(p)\mathcal{F}(s)\mathcal{F}({(H_{nm}^w)^T}) + \mathcal{F}(n)\mathcal{F}({(H_{nm}^w)^T}),
	\end{split}
\end{equation}
where the equation holds due to the convolution theorem.

Mathematically, when the order $(n,m)$ is small, e.g., under a $l$-norm-based constraint condition $\{ (n,m):||(n,m)||_l \le K\}$ with small positive constant $K$, the frequency distribution of $\mathcal{F}({(H_{nm}^w)^T})$ is sparse \cite{ref22}, and $\mathcal{F}({(H_{nm}^w)^T})$ is basically composed of low-frequency coefficients \cite{ref22}.

With this property and the low-pass nature of $p$, we have $\mathcal{F}(p)\mathcal{F}({(H_{nm}^w)^T}) \simeq \mathcal{F}({(H_{nm}^w)^T})$.

As for $\mathcal{F}(n)$, it is random, specifically the phase is completely random and the amplitude is random but constrained by the noise intensity. Hence, for the frequency $({\xi _x},{\xi _y})$ such that $\mathcal{F}({(H_{nm}^w)^T})({\xi _x},{\xi _y}) \simeq 0$, we have $\mathcal{F}(n)({\xi _x},{\xi _y})\mathcal{F}({(H_{nm}^w)^T})({\xi _x},{\xi _y}) \simeq 0$. Note that due to the sparsity of $\mathcal{F}({(H_{nm}^w)^T})$, such $({\xi _x},{\xi _y})$ satisfying $\mathcal{F}({(H_{nm}^w)^T})({\xi _x},{\xi _y}) \simeq 0$ is dominant in the frequency domain. As for other frequency $({\xi _x},{\xi _y})$, the effect of the noise for the frequency coefficients is also mitigated, since the amplitude of $\mathcal{F}({(H_{nm}^w)^T})$ is quite small relative to the amplitude of $\mathcal{F}(n)$.

\renewcommand\thetable{A1} 
\begin{table*}[!t]
	\caption{Definitions of Radial Basis Functions of Unit Disk-based Orthogonal Moments}
	\centering
	\begin{tabular}{cc}
		\toprule
		Method & Radial Basis Function \\ \midrule
		ZM \cite{ref63} &  $R_{nm}^{(\rm{ZM})}(r) = \sqrt {\frac{{n + 1}}{\pi }}  \sum\limits_{k = 0}^{\frac{{n - |m|}}{2}} {\frac{{{{( - 1)}^k}(n - k)!{r^{n - 2k}}}}{{k!(\frac{{n + |m|}}{2} - k)!(\frac{{n - |m|}}{2} - k)!}}}$\\
		PZM \cite{ref64} & $R_{nm}^{(\rm{PZM})}(r) = \sqrt {\frac{{n + 1}}{\pi }} \sum\limits_{k = 0}^{n - |m|} {\frac{{{{( - 1)}^k}(2n + 1 - k)!{r^{n - k}}}}{{k!(n + |m| + 1 - k)!(n - |m| - k)!}}}$ \\ 
		OFMM \cite{ref65} & $R_n^{(\rm{OFMM})}(r) = \sqrt {\frac{{n + 1}}{\pi }}\sum\limits_{k = 0}^n {\frac{{{{( - 1)}^{n + k}}(n + k + 1)!{r^k}}}{{k!(n - k)!(k + 1)!}}}$ \\ 
		CHFM \cite{ref66} & $R_n^{(\rm{CHFM})}(r) = \frac{2}{\pi }{\left( {\frac{{1 - r}}{r}} \right)^{\frac{1}{4}}} \sum\limits_{k = 0}^{\left\lfloor {\frac{n}{2}} \right\rfloor } {\frac{{{{( - 1)}^k}(n - k)!{{(4r - 2)}^{n - 2k}}}}{{k!(n - 2k)!}}}$ \\ 
		PJFM \cite{ref67} & $R_n^{(\rm{PJFM})}(r) =\sqrt {\frac{{(n + 2)(r - {r^2})}}{{\pi (n + 3)(n + 1)}}}\sum\limits_{k = 0}^n {\frac{{{{( - 1)}^{n + k}}(n + k + 3)!{r^k}}}{{k!(n - k)!(k + 2)!}}}$ \\ 
		JFM \cite{ref68} & $R_n^{(\rm{JFM})}(p,q,r) = \sqrt {\frac{{{r^{q - 2}}{{(1 - r)}^{p - q}}(p + 2n) \cdot \Gamma (q + n) \cdot n!}}{{2\pi \Gamma (p + n) \cdot \Gamma (p - q + n + 1)}}}\sum\limits_{k = 0}^n {\frac{{{{( - 1)}^k}\Gamma (p + n + k){r^k}}}{{k!(n - k)!\Gamma (q + k)}}}$ \\ 
		RHFM \cite{ref69} & $	R_n^{(\rm{RHFM})}(r) = \left\{ {\begin{array}{*{20}{c}}
				{\frac{1}{{\sqrt {2\pi r} }}}&{n = 0}\\
				{\sqrt {\frac{1}{{\pi r}}} \sin (\pi (n + 1)r)}&{n > 0\;\& \;n\;{\rm{odd}}}\\
				{\sqrt {\frac{1}{{\pi r}}} \cos (\pi nr)}&{n > 0\;\& \;n\;{\rm{even}}}
		\end{array}} \right.$ \\ 
		EFM \cite{ref70} & $R_n^{(\rm{EFM})}(r) = \frac{1}{{\sqrt {2\pi r} }}\exp (\bm{j}2n\pi r)$ \\ 
		PCET \cite{ref42} & $R_n^{(\rm{PCET})}(r) = \frac{1}{{\sqrt \pi  }}\exp (\bm{j}2n\pi {r^2})$ \\
		PCT \cite{ref42} & $R_n^{(\rm{PCT})}(r) = \left\{ {\begin{array}{*{20}{c}}
				{\frac{1}{{\sqrt \pi  }}}&{n = 0}\\
				{\sqrt {\frac{2}{\pi }} \cos (n\pi {r^2})}&{n > 0}
		\end{array}} \right.$ \\ 
		PST \cite{ref42} & $R_n^{(\rm{PST})}(r) = \sqrt {\frac{2}{\pi }} \sin (n\pi {r^2})$ \\ 
		BFM \cite{ref71} & $	R_n^{(\rm{BFM})}(r) = \frac{1}{{\sqrt {\pi} {{{{J_{v + 1}}({\lambda _n})}}} }}{J_v}({\lambda _n}r)$, ${J_v}(x) = \sum\limits_{k = 0}^\infty  {\frac{{{{( - 1)}^k}}}{{k!\Gamma (v + k + 1)}}} {\left( {\frac{x}{2}} \right)^{v + 2k}}$ \\
		GRHFM \cite{ref72} & $	R_n^{(\rm{GRHFM})}(\alpha,r) = \left\{ {\begin{array}{*{20}{c}}
				{\sqrt{{\frac{\alpha r^{\alpha-2}}{2\pi} }}}&{n = 0}\\
				{\sqrt{{\frac{\alpha r^{\alpha-2}}{\pi} }} \sin (\pi (n + 1)r^\alpha)}&{n > 0\;\& \;n\;{\rm{odd}}}\\
				{\sqrt{{\frac{\alpha r^{\alpha-2}}{\pi} }} \cos (\pi nr^\alpha)}&{n > 0\;\& \;n\;{\rm{even}}}
		\end{array}} \right.$ \\
		GPCET \cite{ref72} & $R_n^{(\rm{GPCET})}(\alpha,r) = \sqrt{{\frac{\alpha r^{\alpha-2}}{2\pi} }}\exp (\bm{j}2n\pi {r^\alpha})$  \\
		GPCT \cite{ref72} & $R_n^{(\rm{GPCT})}(\alpha,r) = \left\{ {\begin{array}{*{20}{c}}
				{\sqrt{{\frac{\alpha r^{\alpha-2}}{2\pi} }}}&{n = 0}\\
				{\sqrt{{\frac{\alpha r^{\alpha-2}}{\pi} }} \cos (\pi nr^\alpha)}&{n > 0}
		\end{array}} \right.$  \\
		GPST \cite{ref72} &  $R_n^{(\rm{GPST})}(\alpha,r) = \sqrt{{\frac{\alpha r^{\alpha-2}}{\pi} }} \sin (n\pi {r^\alpha})$ \\
		FJFM \cite{ref40} &  $R_n^{(\rm{FJFM})}(\alpha,p,q,r) = \sqrt {\frac{{\alpha {r^{\alpha q - 2}}{{(1 - {r^\alpha })}^{p - q}}(p + 2n) \cdot \Gamma (q + n)n!}}{{2\pi \Gamma (p + n) \cdot \Gamma (p - q + n + 1)}}} \sum\limits_{k = 0}^n {\frac{{{{( - 1)}^k}\Gamma (p + n + k){r^{\alpha k}}}}{{k!(n - k)!\Gamma (q + k)}}} $ \\
		\bottomrule
	\end{tabular}
\end{table*}

\renewcommand\thetable{A2}   
\begin{table*}[!t]
	\caption{Precision, Recall, and F1 Scores (\%) for Different DIR Settings on the FAU Copy-Move Forensic Benchmark.}
	\centering
	\begin{tabular}{ccccccccccc}
		\toprule
		Parameter & Setting & $\#$1   & $\#$2   & $\#$3   & $\#$4   & $\#$5   & $\#$6   & $\#$7   & $\#$8   & $\#$9   \\
		\midrule
		\multirow{3}{*}{$K$}      & 3          & \checkmark     &       &       & \checkmark     & \checkmark     & \checkmark     & \checkmark     & \checkmark     & \checkmark     \\
		& 1          &       & \checkmark     &       &       &       &       &       &       &       \\
		& 5          &       &       & \checkmark     &       &       &       &       &       &       \\
		\midrule
		\multirow{3}{*}{$w$}      & 8 $\sim$ 32  & \checkmark     & \checkmark     & \checkmark     &       &       & \checkmark     & \checkmark     & \checkmark     & \checkmark     \\
		& 10 $\sim$ 20 &       &       &       & \checkmark     &       &       &       &       &       \\
		& 6 $\sim$ 36  &       &       &       &       & \checkmark     &       &       &       &       \\
		\midrule
		\multirow{5}{*}{$R$}    & PCT        & \checkmark     & \checkmark     & \checkmark     & \checkmark     & \checkmark     &       &       &       &       \\
		& PCET       &       &       &       &       &       & \checkmark     &       &       &       \\
		& ZM         &       &       &       &       &       &       & \checkmark     &       &       \\
		& OFMM       &       &       &       &       &       &       &       & \checkmark     &       \\
		& BFM        &       &       &       &       &       &       &       &       & \checkmark     \\
		\midrule\midrule
		\multirow{3}{*}{Scores} & Precision  & 96.75 & 96.75 & 96.75 & 97.95 & 96.94 & 96.50 & 96.65 & 96.68 & 92.80 \\
		& Recall     & 92.44 & 89.38 & 92.67 & 93.69 & 91.25 & 91.00 & 90.43 & 92.18 & 89.58 \\
		& F1         & 93.62 & 92.41 & 93.67 & 95.18 & 93.09 & 92.42 & 92.14 & 93.41 & 90.39\\
		\bottomrule
	\end{tabular}
\end{table*}

\renewcommand\thefigure{A1}  
\begin{figure*}[!t]
	\centering
	\includegraphics[scale=1]{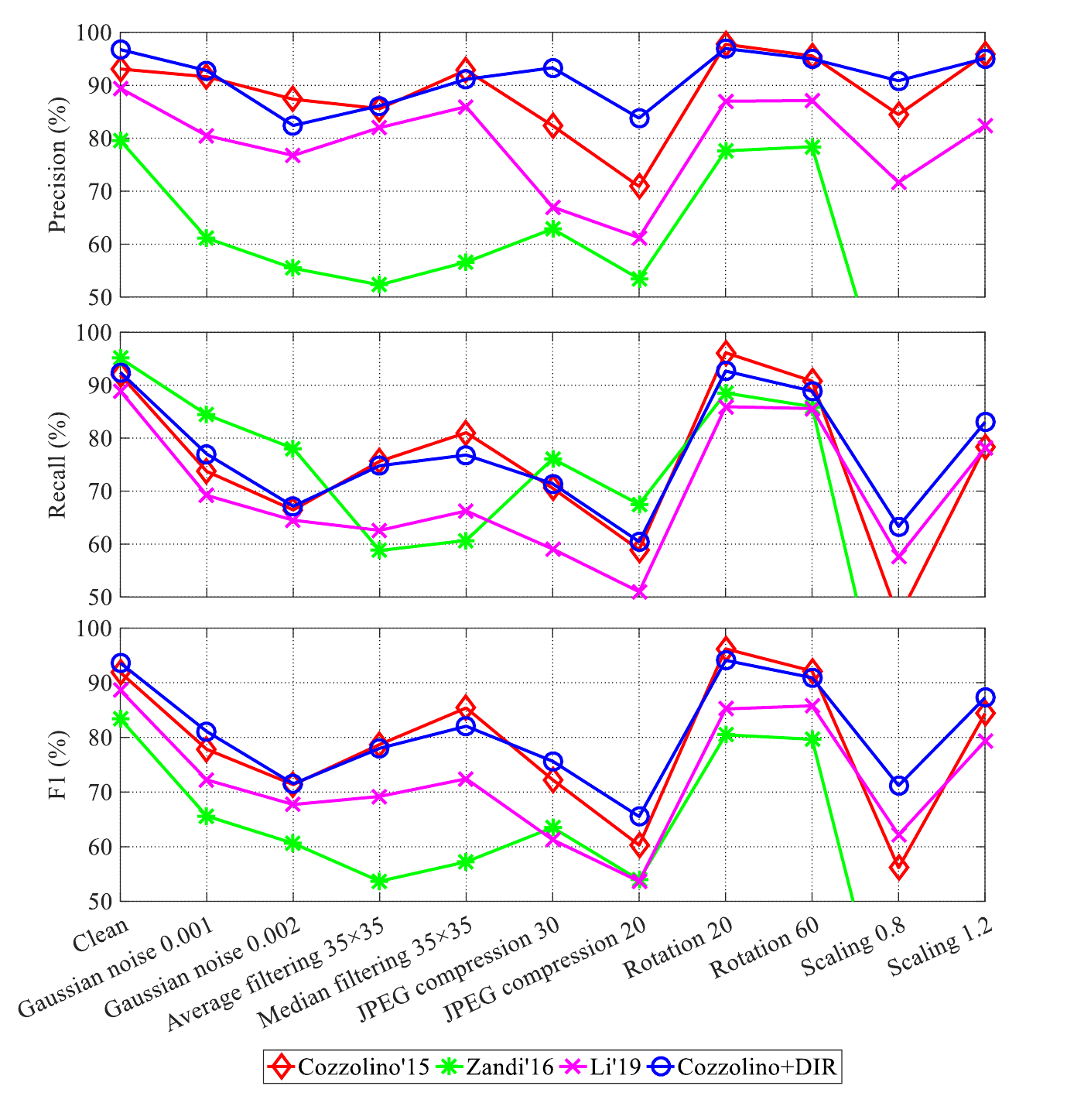}
	\centering
	\caption{Precision, recall, and F1 curves by different copy-move forgery detection methods in comprehensive robustness experiment.}
\end{figure*}

\renewcommand\thetable{A3}   
\begin{table*}[!t]
	\caption{Precision, Recall, and F1 Scores (\%) for Different DIR Settings in Hash Robustness Experiment.}
	\centering
	\begin{tabular}{ccccccccccc}
		\toprule
		Parameter & Setting & $\#$1   & $\#$2   & $\#$3   & $\#$4   & $\#$5   & $\#$6   & $\#$7   & $\#$8   & $\#$9   \\
		\midrule
		\multirow{3}{*}{$K$}      & 3          & \checkmark     &       &       & \checkmark     & \checkmark     & \checkmark     & \checkmark     & \checkmark     & \checkmark     \\
		& 1          &       & \checkmark     &       &       &       &       &       &       &       \\
		& 5          &       &       & \checkmark     &       &       &       &       &       &       \\
		\midrule
		\multirow{3}{*}{$w$}      & 8 $\sim$ 12  & \checkmark     & \checkmark     & \checkmark     &       &       & \checkmark     & \checkmark     & \checkmark     & \checkmark     \\
		& 10 &       &       &       & \checkmark     &       &       &       &       &       \\
		& 6 $\sim$ 16  &       &       &       &       & \checkmark     &       &       &       &       \\
		\midrule
		\multirow{5}{*}{$R$}    & PCT        & \checkmark     & \checkmark     & \checkmark     & \checkmark     & \checkmark     &       &       &       &       \\
		& PCET       &       &       &       &       &       & \checkmark     &       &       &       \\
		& ZM         &       &       &       &       &       &       & \checkmark     &       &       \\
		& OFMM       &       &       &       &       &       &       &       & \checkmark     &       \\
		& BFM        &       &       &       &       &       &       &       &       & \checkmark     \\
		\midrule\midrule
		\multirow{3}{*}{Scores} & Precision & 80.72 & 84.11 & 79.19 & 79.34 & 81.29 & 82.13 & 85.68 & 80.63 & 80.44 \\
		& Recall    & 69.35 & 68.31 & 69.53 & 69.80 & 69.30 & 68.78 & 67.15 & 69.40 & 69.12 \\
		& F1        & 69.94 & 71.75 & 68.83 & 69.14 & 70.54 & 70.51 & 71.97 & 69.93 & 69.47\\
		\bottomrule
	\end{tabular}
\end{table*}

Based on the above assumptions, it is reasonable to derive the following approximation:
\begin{equation}\nonumber
	\begin{split}
		& \mathcal{F}( \left< f,V_{nm}^{uvw} \right> )\\
		& = \mathcal{F}(p)\mathcal{F}(s)\mathcal{F}({(H_{nm}^w)^T}) + \mathcal{F}(n)\mathcal{F}({(H_{nm}^w)^T})\\
		& \simeq \mathcal{F}(s)\mathcal{F}({(H_{nm}^w)^T})\\
		& = \mathcal{F}( \left< s,V_{nm}^{uvw} \right> ),
	\end{split}
\end{equation}
i.e., the DIR-based low-order moments for $f$ and $s$,  $\left< f,V_{nm}^{uvw} \right>$ and  $\left< s,V_{nm}^{uvw} \right>$, are approximation in frequency domain, meaning certain robustness in many practical tasks for the signal corruptions like blurring and noise.

\section{Complementary Experiments for Copy-Move Forgery Detection}

We have conducted several complementary experiments for copy-move forgery detection w.r.t. different parameter settings, different basis function definitions, and systematic image attacks.

In Table A2, the copy-move forensic accuracy on the FAU benchmark under different DIR settings is provided, involving three different values of $K$/$w$ and five different radial basis function definitions. As can be seen from this table, the forensic localization scores fluctuate slightly (in a reasonable interval) as the changes of settings, and the setting used in the paper (i.e., setting $\#$1) is not a special setting with highest scores. These phenomena imply that the DIR-based copy-move algorithm is not sensitive to such settings and the relevant experimental evaluations with setting $\#$1 in the paper are convincing (i.e., not an isolated case).

In Fig. A1, the forensic accuracy curves on the FAU benchmark under different attacks are provided, involving six signal corruptions of whole images and four geometric transformations of manipulated regions. As can be seen from Fig. A1, for signal corruptions, the proposed method exhibits very similar scores to Cozzolino’15, which is in line with expectations since Cozzolino’15 is actually based on a single-scale DIR, while the other competing methods (i.e., Zandi’16 and Li’19) exhibit lower scores under most attacks. For geometric transformations, Cozzolino’15 and its DIR version are also systematically more robust than other competing methods. Rotation attack does not lead to significant differences between the Cozzolino’15 and its DIR version, implying that the rotation invariance is maintained in our multi-scale DIR framework. Under the scaling, especially with scale factor 0.8, the proposed DIR version significantly outperforms Cozzolino’15 (also Zandi’16 and Li’19), which is consistent with our original intention of introducing multi-scale DIR.

\section{Complementary Experiments for Perceptual Hashing}

We have also conducted several complementary experiments for perceptual hashing w.r.t. different parameter settings and different basis function definitions.

In Table A3, the perceptual-hashing based forensic accuracy on the RTD benchmark under different DIR settings is provided, involving three different values of $K$/$w$ and five different radial basis function definitions. Note that the scores in Table A4 are averaged over all the attacks listed in Table 9. Here, a very similar conclusion can be observed, i.e., the DIR-based perceptual hashing algorithm is not sensitive to such settings and the relevant experimental evaluations with setting $\#$1 in the paper are convincing.

\section{A Remark on the Forensic Applicability}

Following the taxonomy of DARPA, digital media forensic should look for digital integrity, physical integrity, and semantic integrity \cite{ref1}. From the representation perspective, such digital, physical, and semantic integrities mainly correspond to the high-order, mid-order, and low-order statistics of the image, respectively \cite{ref1}. As an image orthogonal analysis tool, our DIR has the beneficial property of information preservation (the whole image information is preserved in orthogonal moment space) \cite{ref22}, i.e., the full-order statistics provided by DIR allow a uniform discriminability for broad forensic scenarios.

We would like to clearly state that the forensic tasks listed in this paper, i.e., copy-move detection and perceptual hashing by semantic comparison, both belong to semantic-level algorithm. However, the discriminability provided by DIR is in fact not limited to such forensic tasks. We have initially explored the application of DIR to digital-level forensic tasks, i.e., splicing and inpainting detection by exposing digital artifacts, based on the coefficient distribution (kurtosis) modeling.

\ifCLASSOPTIONcaptionsoff
  \newpage
\fi



%
\bibliographystyle{IEEEtran}
\bibliography{paper}



%

\begin{IEEEbiography}[{\includegraphics[width=1in,height=1.25in,clip,keepaspectratio]{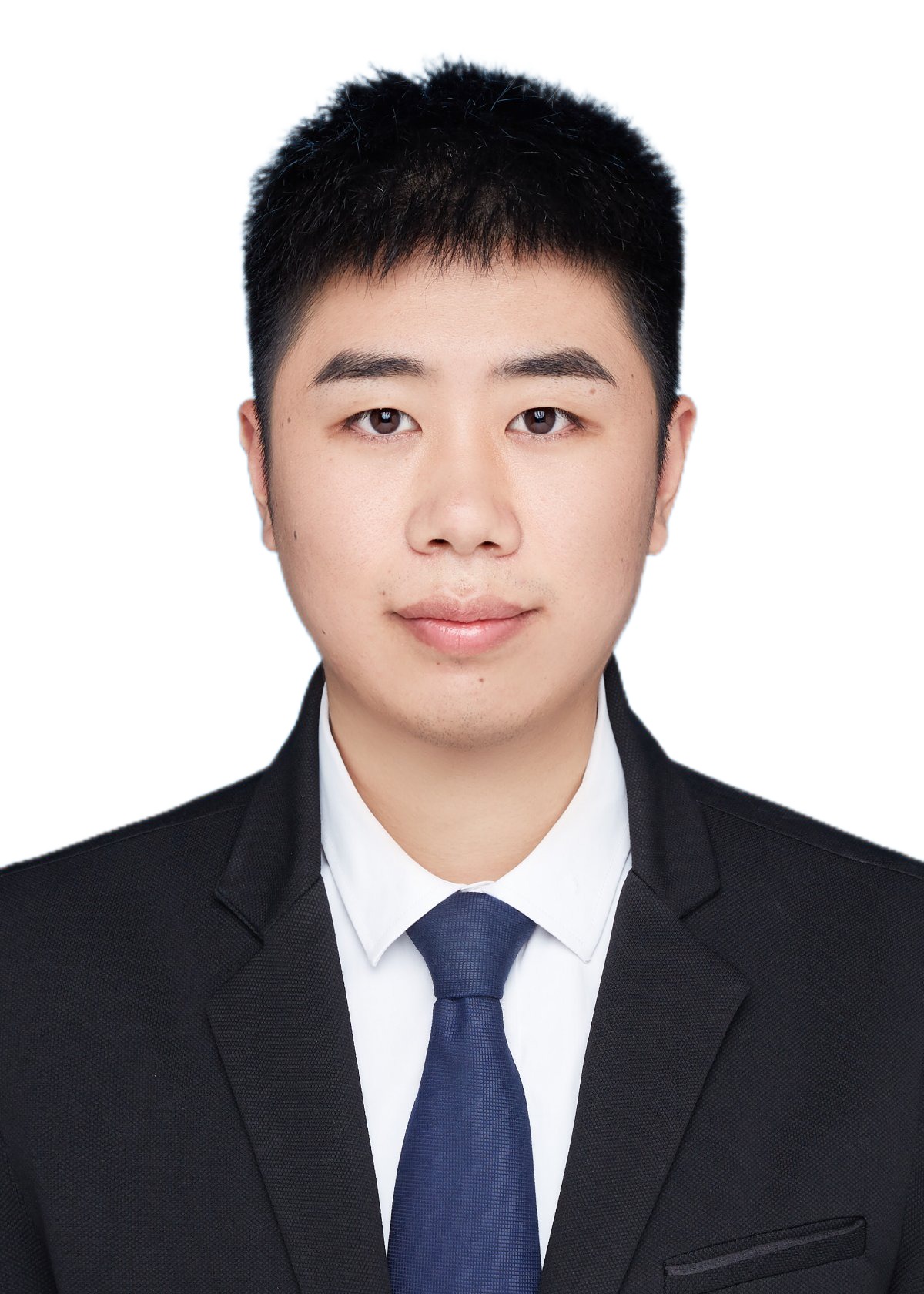}}]{Shuren Qi}
	received the B.A. and M.E. degrees from Liaoning Normal University, Dalian, China, in 2017 and 2020 respectively. He is currently pursuing the Ph.D. degree in computer science at Nanjing University of Aeronautics and Astronautics, Nanjing, China. His research interests include invariant feature extraction and visual signal representation with applications in robust pattern recognition and multimedia forensics/security.
\end{IEEEbiography}
\begin{IEEEbiography}[{\includegraphics[width=1in,height=1.25in,clip,keepaspectratio]{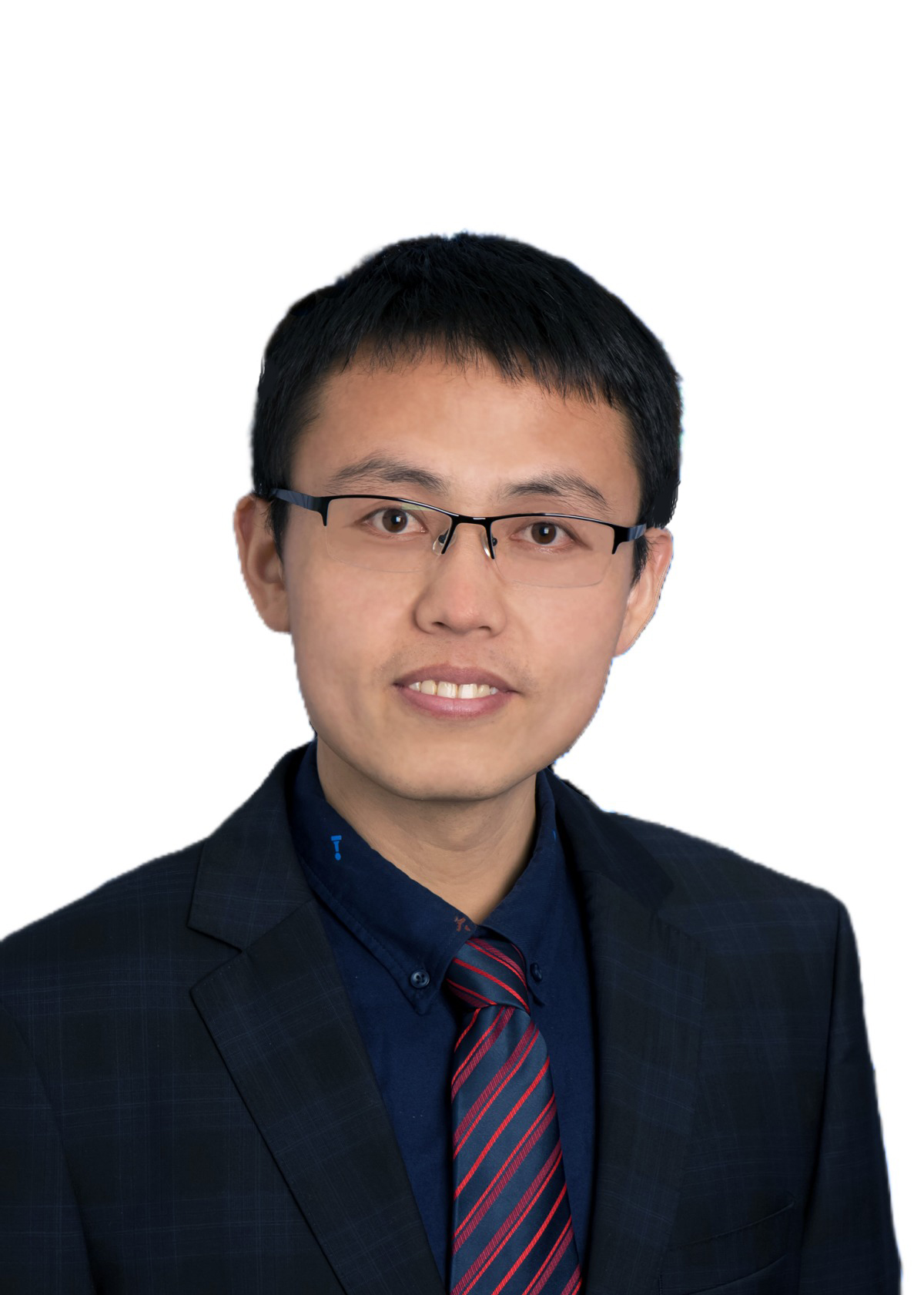}}]{Yushu Zhang}
	(Member, IEEE) received the Ph.D. degree in computer science from Chongqing University, Chongqing, China, in 2014. He held various research positions with the City University of Hong Kong, Southwest University, University of Macau, and Deakin University. He is currently a Professor with the College of Computer Science and Technology, Nanjing University of Aeronautics and Astronautics, Nanjing, China. His research interests include multimedia processing and security, artificial intelligence, and blockchain. Dr. Zhang is an Associate Editor of \emph{Signal Processing} and \emph{Information Sciences}.
\end{IEEEbiography}

\begin{IEEEbiography}[{\includegraphics[width=1in,height=1.25in,clip,keepaspectratio]{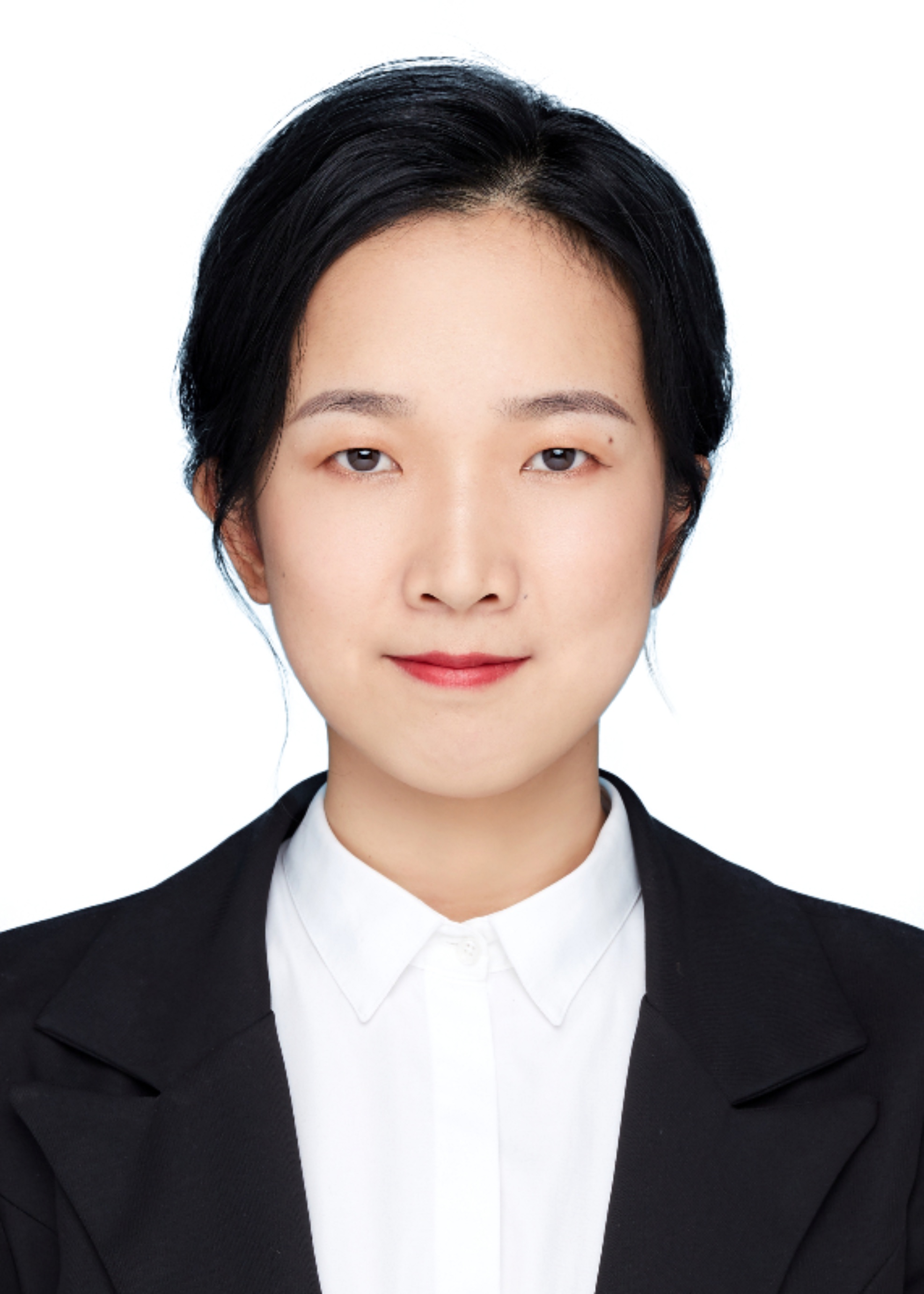}}]{Chao Wang}
	received the B.S. and M.S. degrees from Liaoning Normal University, Dalian, China, in 2017 and 2020 respectively. She is currently pursuing the Ph.D. degree in computer science at Nanjing University of Aeronautics and Astronautics, Nanjing, China. Her research interests include trustworthy artificial intelligence, adversarial learning, and media forensics.
\end{IEEEbiography}

\begin{IEEEbiography}[{\includegraphics[width=1in,height=1.25in,clip,keepaspectratio]{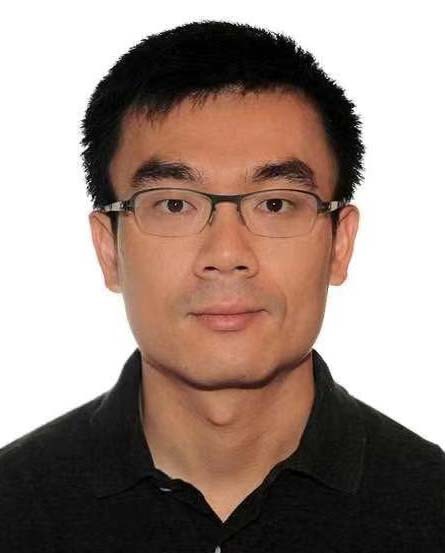}}]{Jiantao Zhou}
	 (Senior Member, IEEE) received the B.E. degree from the Department of Electronic Engineering, Dalian University of Technology, in 2002, the M.E. degree from the Department of Radio Engineering, Southeast University, in 2005, and the Ph.D. degree from the Department of Electronic and Computer Engineering, The Hong Kong University of Science and Technology, in 2009. He held various research positions with the University of Illinois at Urbana-Champaign, The Hong Kong University of Science and Technology, and McMaster University. He is currently an Associate Professor with the Department of Computer and Information Science, Faculty of Science and Technology, University of Macau. He holds four granted U.S. patents and two granted Chinese patents. His research interests include multimedia security and forensics, and multimedia signal processing. He has coauthored three articles that received the Best Paper Award from the \emph{IEEE Pacific-Rim Conference on Multimedia} in 2007, the Best Student Paper Award, and the Best Paper Runner Up Award from the \emph{IEEE International Conference on Multimedia and Expo} in 2016 and 2020. He has been an Associate Editor of \emph{IEEE Transactions on Image Processing} since 2019.
\end{IEEEbiography}

\begin{IEEEbiography}[{\includegraphics[width=1in,height=1.25in,clip,keepaspectratio]{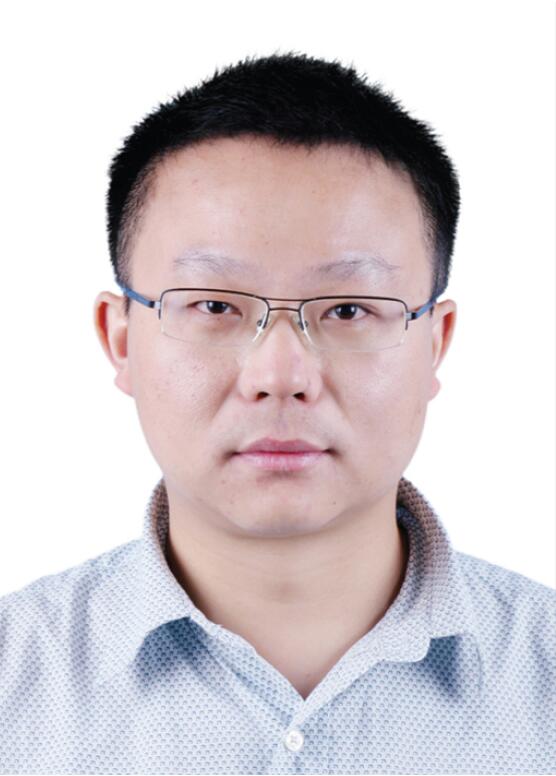}}]{Xiaochun Cao}
	(Senior Member, IEEE) received the B.E. and M.E. degrees in computer science from Beihang University, Beijing, China, and the Ph.D. degree in computer science from the University of Central Florida, Orlando, FL, USA. After graduation, he spent about three years at ObjectVideo Inc. as a Research Scientist. From 2008 to 2012, he was a Professor with Tianjin University, Tianjin, China. He is currently a Professor with School of Cyber Science and Technology, Shenzhen Campus, Sun Yat-sen University, Shenzhen, China. He is a Fellow of the IET. His dissertation was nominated for the University of Central Florida’s university-level Outstanding Dissertation Award. In 2004 and 2010, he was the recipient of the Piero Zamperoni Best Student Paper Award at the \emph{International Conference on Pattern Recognition}. He is on the Editorial Board of the \emph{IEEE Transactions on Image Processing}, \emph{IEEE Transactions on Multimedia}, and \emph{IEEE Transactions on Circuits and Systems for Video Technology}.
\end{IEEEbiography}




\end{document}